\documentclass[USenglish,oneside,twocolumn]{article}

\usepackage{tabularx}
\usepackage{pgfplots}
\usepackage[title]{appendix}
\usepackage[utf8]{inputenc}
\usepackage[big]{dgruyter_NEW}
\usepackage{cleveref}

\cclogo{\includegraphics{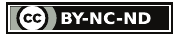}}

\begin{document}

\author*[1]{Edwin Dauber}
\author[2]{Aylin Caliskan}
\author[3]{Richard Harang}
\author[4]{Gregory Shearer}
\author[5]{Michael Weisman}
\author[6]{Frederica Nelson}
\author[7]{Rachel Greenstadt}
\affil[1]{Drexel University, Email: egd34@drexel.edu}
\affil[2]{George Washington University, Email: aylin@email.gwu.edu}
\affil[3]{Sophos Data Science Team, Email: rich.harang@sophos.com}
\affil[4]{ICF International, Email: gregory.g.shearer.ctr@mail.mil}
\affil[5]{United States Army Research Laboratory, Email: michael.j.weisman2.civ@mail.mil}
\affil[6]{United States Army Research Laboratory, Email: frederica.f.nelson.civ@mail.mil}
\affil[7]{New York University, Email: greenstadt@nyu.edu}

\title{\huge{Git Blame Who?: Stylistic Authorship Attribution of Small, Incomplete Source Code Fragments}}
\runningtitle{Git Blame Who?: Stylistic Authorship Attribution of Small, Incomplete Source Code Fragments}

\begin{abstract}
{Program authorship attribution has implications for the privacy of programmers who wish to contribute code anonymously. While previous work has shown that individually authored complete files can be attributed, these efforts have focused on such ideal data sets as contest submissions and student assignments. We explore the problem of authorship attribution ``in the wild,'' examining source code obtained from open-source version control systems, and investigate how contributions can be attributed to their authors, either on an individual or a per-account basis.  In this work, we present a study of attribution of code collected from collaborative environments and identify factors which make attribution of code fragments more or less successful. 
For individual contributions, we show that previous methods (adapted to be applied to short code fragments) yield an accuracy of approximately 50\% or 60\%, depending on whether we average by sample or by author, at identifying the correct author out of a set of 104 programmers.  By ensembling the classification probabilities of a sufficiently large set of samples belonging to the same author we achieve much higher accuracy for assigning the set of samples to the correct author from a known suspect set. Additionally, we propose the use of \emph{calibration curves} to identify which samples are by unknown and previously unencountered authors.} 
\end{abstract}

\keywords{Stylometry, code authorship attribution}

  \journalname{Proceedings on Privacy Enhancing Technologies}
\DOI{Editor to enter DOI}
  \startpage{1}
  \received{..}
  \revised{..}
  \accepted{..}

  \journalyear{..}
  \journalvolume{..}
  \journalissue{..}

\maketitle

\section{Introduction}

The attribution of source code, be it in the form of small fragments such as version control system commits to an open source project or sequential differences in source code samples, entire single-author project files, or multiple small fragments all linked to a common account, has numerous potential applications, from employer-mandated non-compete clauses to the (potential) attribution of the authors of malicious code.  While the problem of attributing source code samples known to be written by a single author has been examined in some depth using the Google Code Jam dataset \cite{gcj,burrows2010source}, extending these techniques to real-world problems -- such as deanonymizing the account of a programmer who has contributed to an open source project -- is much harder, and has received less attention.  Indeed, prior to this work it was an open question as to whether such attribution could be done in a realistic setting\footnote{``I will believe that code stylometry works when it can be shown to work on big github commit histories instead of GCJ dataset'' by Halvar Flake on Twitter: \url{https://twitter.com/halvarflake/status/682263306095181824}}.

The ability to attribute source code in collaborative settings has serious privacy implications.  Many open-source software projects are collaborative, so being able to attribute collaborative code increases the chances of deanonymizing pseudonymous open-source contributors.  For some, such as activists working on censorship circumvention, this may pose a very real danger.  On the other side, a lot of malicious software is either written collaboratively or evolved over time, and much code in the corporate world, where copyright or plagiarism concerns are more likely to arise, is also collaborative.

Collaboratively written source code (or any source code which has passed through the hands of multiple programmers) introduces several novel complications to the problem of attribution that are not present in previous work focused on single files.  Files may have a primary author who has contributed the majority of the code, with other authors making relatively inconsequential additions.  A block of code may be written by one author and later modified by another, perhaps merging their respective styles.  Individual author contributions may be scattered across multiple files and consist of fragments as small as a single line of source code. Using real world source code collected from the online collaborative platform GitHub \cite{git}, we show that these problems can be overcome, or at least mitigated, particularly when multiple modifications can be linked to a single unknown identity (for example, a single pseudonymous version control system account).

More concretely, using the example of a company which wishes to identify an employee who has contributed to an open source project -- perhaps in violation of a non-compete or intellectual property clause -- we assume a known set of suspect programmers (such as the employees of the company) and some form of segmentation and grouping by authorship of the code that the company wishes to attribute (such as accounts on a version control system hosting the code in question).  Within this setting, we present an extension to the technique used by Caliskan-Islam et al.\  which performs stylistic authorship attribution of a collection of partial source code samples written by the same unknown programmer with high accuracy for a set of 104 suspect programmers \cite{caliskan2015anonymizing}.  We note that the samples we work with can be as small as a single line of code, and so can contain very little information with which to determine authorship.  By contrast, the unmodified technique used by Caliskan-Islam et al.\ technique for source code attribution achieves an individual accuracy on individual samples of approximately 50\% or 60\% under standard circumstances, depending on whether we count accuracy by sample or by author. 

To address the open world problem in which the true programmer could be someone outside of the suspect set (in our working example, perhaps no employee has actually contributed to the project in question) we examine a technique using the classifier's output probability, or \emph{confidence}. We construct \emph{calibration curves} to indicate the accuracy for collections which were attributed with given confidence, and analysts can use these curves to set a threshold below which to more carefully examine authorship due to higher probability of being outside of the suspect set \cite{niculescu2005predicting}.
We then proceed to show how this calibration curve can be used at the level of individual samples to mitigate the cost of mis-attribution in the absence of grouping of samples. In this scenario, the threshold will need to be set higher to catch most mis-attributions, and will cause a larger percentage of correctly attributed samples to require review.

We demonstrate attribution under various scenarios. The easiest scenario, which we refer to as \emph{multiple sample attribution}, features the \emph{closed world assumption}, under which the author is among our suspects, and the \emph{version control assumption}, under which we assume that the multi-authored code file is segmented and those segments are grouped by (unknown) author.
We also remove the closed world assumption and address the open world version of multiple sample attribution. We also show results with a relaxation of the version control assumption which only assumes segmentation, and not grouping, for comparison purposes and to demonstrate the usefulness of the calibration curve. We refer to this variant of the problem as \emph{single sample attribution}, and observe it both under the closed world assumption and in the open world.

\subsection{Contributions}
We present results of the first evaluation of the Caliskan-Islam et al.\ technique under the conditions of collaborative code, which to our knowledge is also the first evaluation of any stylometric technique on collaborative code \cite{caliskan2015anonymizing}. Specifically, we use git blame to segment collaborative code files into pieces which we can assign to a singular programmer, and show that, unmodified except for excluding feature reduction, the Caliskan-Islam et al.\ approach performs much worse on such segments (achieving accuracy of 50\% or 60\% as opposed to the over 98\% accuracy expected for whole files based on their initial work). We also perform attributability analysis, to determine what factors may make samples easier or harder to attribute.

We also present two modifications to the Caliskan-Islam et al.\ approach which better handle such difficult conditions. First, we show that a form of ensembling in which we ensemble outputs of multiple linked samples for the same classifier, rather than by ensembling the outputs of different classifiers on the same sample, can greatly improve accuracy in the event that we are able to link several samples to the same unknown programmer.  Second, we show that we can use \emph{calibration curves} to determine the trustworthiness of our attributions for a given sample or set of samples.

\section{Related Work}
\label{sec:related}

We note two broad, primary categories of related work.  The most similar is past work in source code authorship attribution, largely focusing on single files.  We also examine work in the area of plain-text authorship attribution.  While the two domains have evolved different feature sets and classification techniques, recent work in text authorship attribution is related to our work.

\subsection{Source Code Authorship Attribution}

An important overall note about related work is that we are the first to attempt attribution of short, incomplete, and typically uncompilable code samples.  To our knowledge, all past attempts at source code authorship attribution have worked with complete code samples which solve some problem, while we work with small building blocks of code which in almost all cases cannot be compiled when separated from their surrounding code.

The primary piece of prior research related to our work is the work of Caliskan-Islam et al.\ using random forests to perform authorship attribution of Google Code Jam submissions, with features extracted from the abstract syntax tree (AST) \cite{caliskan2015anonymizing}.  We begin with their feature set and classification method as the basis for our work, but are forced to significantly modify their analysis to be able to handle small pieces of code that have been attributed to individual authors by use of git blame, rather than complete source code files. As a result of focusing on small segments of code, our feature vectors are much more sparse and we are unable to prune the feature set through use of information gain as they did.  Their work also looks at the open world problem, and uses classification confidence to set a threshold below which to reject classifications, which we further develop to both address the open world problem and the \emph{single sample attribution} problem.  They attribute data from Google Code Jam, which as a programming competition creates laboratory conditions for the data, while we attribute data from GitHub, giving us real world conditions, including many of the issues presented in the introduction, and ``ground truth'' labeling problems introduced by the use of git blame.  We demonstrate that their techniques can be adapted to handle more difficult attribution tasks which are of interest in real world situations and have not been previously examined.

Abuhamad et al.\ proposed a system called DL-CAIS which highly effectively attributes code at large scale and cross-language \cite{abuhamad2018large}. However, while they do test on code from GitHub, their work does not address the problem of collaborative code, nor the problem of small code samples, and their deep learning approach to learning feature representations should be less effective for smaller samples.

There are many other proposed methods and feature sets for source code de-anonymization, but most of those methods had worse accuracy and smaller suspect sets.  Therefore, while combining these techniques may allow us to boost accuracy, for the purposes of this work we do not consider them further.  Frantzeskou et al.\ used byte level n-grams to achieve high accuracy with small suspect sets \cite{frantzeskou2008examining,frantzeskou2007identifying,frantzeskou2006effective}.  The use of ASTs for authorship attribution was pioneered by Pellin and used on pairs of Java programs in order to ensure that the studied programs had the same functionality \cite{pellin2000using}.  Ding and Samadzadeh studied a set of 46 programmers and Java using statistical methods \cite{ding2004extraction}.  MacDonnel et al.\ analyzed C++ code from a set of 7 professional programmers using neural networks, multiple discriminant analysis, and case-based reasoning \cite{macdonell1999software}.  Burrows et al.\ proposed techniques that achieved high accuracy for small suspect sets, but had poor scalability \cite{burrows2010source,burrows2007source,burrows2009application}

Spafford and Weeber were among the first to suggest performing authorship attribution on source code \cite{spafford1993software}.  However, while they proposed some features to do so, they did not propose an automated method nor a case study.  Hayes and Offutt performed a manual statistical analysis of 5 professional programmers and 15 graduate students, and found that programmers do have distinguishable styles which they use consistently \cite{hayes2010recognizing}.

In no case that we are aware of did any of the previous authors investigate the various factors that make attribution more or less difficult via machine learning techniques.  

For ground truth we use git blame to assign authorship to individual lines of code.  Git blame is a heuristic which attributes code to the last author to modify that code.  Meng et al.\ proposed a tool called git-author which assigns weighted values to contributions in order to better represent the evolution of a line of code \cite{meng2013mining}.  This tool creates a \emph{repository graph} with commits as nodes and development dependencies as edges, and then defines \emph{structural authorship} and \emph{weighted authorship} for each line of code, where \emph{structural authorship} is a subgraph of the repository graph with respect to that line of code and \emph{weighted authorship} is a vector of programmer contributions derived from structural authorship.

\subsection{Text Authorship Attribution}

The primary piece of related research in the domain of text authorship attribution is the work by Overdorf and Greenstadt in cross-domain authorship attribution \cite{overdorf2016blogs}.  This work links authorship between blogs, tweets, and Reddit comments.  This work is related to ours in two primary ways.  First, and most obviously, they work with short text in the forms of tweets and Reddit comments.  For these domains, they use a technique of merging text before extracting features.  
More significantly, they also use a method of averaging the probabilities for multiple samples to classify collections of samples.  We demonstrate that this technique is similarly applicable in the source code domain, and that we get excellent results even with averaging a small number of samples.  However, they make no effort to classify individual tweets, while we successfully classify samples of code as short as a few lines.

To address the open world problem,  Stolerman et al.\ introduce a method called \emph{Classify-Verify} which augments classification with \emph{verification} \cite{stolerman2013classify}.  Authorship verification is the problem of determining if a document $D$ was written by an author $A$.  Among the verification methods they consider is classifier confidence.  We apply the same intuition to source code to determine which classifications to trust and which to reject.

\section{Methodology}
\label{sec:method}

\begin{figure*}[htb]
\centering
\includegraphics[width=\textwidth]{gitblamewhomethod.png}
\caption{Our method for authorship attribution of partial source code samples has multiple preprocessing and postprocessing stages.}
\label{fig:workflow}
\end{figure*}

Our method is summarized in Figure \ref{fig:workflow}. We begin by collecting C++ repositories on GitHub, and then breaking collaborative files from those repositories into smaller pieces using git blame, as detailed in Section \ref{dataPrep}. For each individual piece of code, we extract the AST and use that to extract a feature vector, as described in Section \ref{features}. We then proceed to perform attribution of each sample using a random forest as described in Section \ref{ssa}. Then we average the classifier probability of linked samples as described in Section \ref{msa} and construct a calibration curve as described in Section \ref{cc}. In Appendix \ref{sec:example} we provide an example walkthrough of the method for a selection of code samples, for those interested.

\subsection{Problem Statement}

In this paper we take on the role of an analyst attempting to attribute source code and break anonymity. The analyst has the job of preparing a report indicating the conclusions reached through the attribution process.  This may indicate an author, likelihood of an author, a subset of the suspect set to investigate further, or likelihood of the true author not being in the suspect set, as the data requires.

We assume that the collaboratively written code we are examining has been pre-segmented by author. This segmentation may be from version control commits, git blame, or some other method of decomposition; we only assume that we have small samples which can be reasonably attributed to a single individual. We also assume that we have training data which consists of similarly segmented code samples by our suspect programmers, rather than full files. Note that this later assumption does not particularly limit us in practical application because we can artificially segment a single authored file if necessary.

In our primary case, we assume that we have multiple \emph{linked} samples by the same (unknown) author, which we call the \emph{version control assumption}.  For example, we may have multiple code samples corresponding to an account on a version control system. We refer to this case as \emph{multiple sample attribution} or as \emph{account attribution}.

Formally, we have a set of source code samples $D$  written by an unknown author $A$ and a set of $n$ suspects $S=\{A_1 \ldots A_n\}$, and for each suspect $A_i$ we have a set of samples $D_i$.  Our goal is to correctly attribute $D$ to an $A_i$.  Where not otherwise stated, we assume the \emph{closed world}, in which one of the $n$ suspects is the true author.

While we believe that most forms of segmentation naturally lead to linking, we acknowledge that by presenting a technique based on linking we may create the assumption that to defend against it one only needs to contribute in a way which prevents linking, such as through guest ``accounts'' or throwaway accounts. Therefore, we not only evaluate the baseline where the cardinality of our code sample sets is 1, but also present ways analysts can interpret the results to compensate for the lower accuracy. We refer to this scenario as \emph{single sample attribution}. Thus, while we show that the version control assumption makes attribution easier and more accurate, it is not a prerequisite for attribution. 

\subsection{Problem Model}
\label{tm}

In this paper we assume an analyst with access to a corpus of source code belonging to each of their suspects or persons of interest. This corpus could come from an employer's system, educational records, a government database, or publicly claimed code online.  Given the large amount of publicly claimed code available from programmers belonging to various potential communities of interest, this assumption is not unrealistic.  We assume anonymous programmers who have contributed to a collaborative code file, but have not taken measures to conceal their coding style.  We note that in a real-world attribution task, stylometric techniques would not be used in a vacuum, but in conjunction with other investigative tools and with the final attribution at the judgment of a human with access to more information than just a final prediction.  For the sake of this paper we assume the role of an analyst limited to stylometric techniques, with outside knowledge limited to a pre-determined suspect set and possible linkage between samples belonging to the same version control account.
The role of the analyst is to use the available information to identify an author along with a measure of how likely the identification is to be correct.  
In order to do this, the analyst will not only need to run the machine learning algorithm, but also analyze the output, and will construct curves to assist in interpreting the results.  

\subsection{Data Preparation}
\label{dataPrep}

We collected data from public C++ repositories on GitHub.  We collected repositories which list C++ as the primary language, starting from 14 seed contributors and spidering through their collaborators.  Doing this, we collected data from 1649 repositories and 1178 programmers, although in future processing steps we found that many of these programmers and repositories had insufficient data.  Additionally, some of these programmers were renamed or group accounts, while some repositories included text other than C++ code which had to be discounted.  After eliminating those, completing data processing, and setting the threshold to at least 150 samples per author with at least 1 line of code (excluding whitespace), we were left with 104 programmers.  We note that this threshold was chosen with an experimental mindset to ensure that we had sufficient data for both training and testing sets.

We note that unlike in the work of Caliskan-Islam et al., we do not strip comments \cite{caliskan2015anonymizing}.  While comments can be problematic due to the ease with which they can be used for obfuscation or imitation and the possibility of over-specificity, because we are working in an overall low information environment we judge that using all information available within the text of the code sample is necessary to compensate.

We used git blame on each line of code, and for each set of consecutive lines blamed to the same programmer we encapsulated those lines in a dummy main function and extracted features from the AST as in the work of Caliskan-Islam et al.~\cite{caliskan2015anonymizing}.  However, unlike in their work we cannot use information gain to prune the feature set due to having extremely sparse feature vectors and therefore few features with individual information gain.

\begin{figure}[htb]
\centering
\includegraphics[trim={1cm 7cm 1cm 7cm},clip, width=\columnwidth]{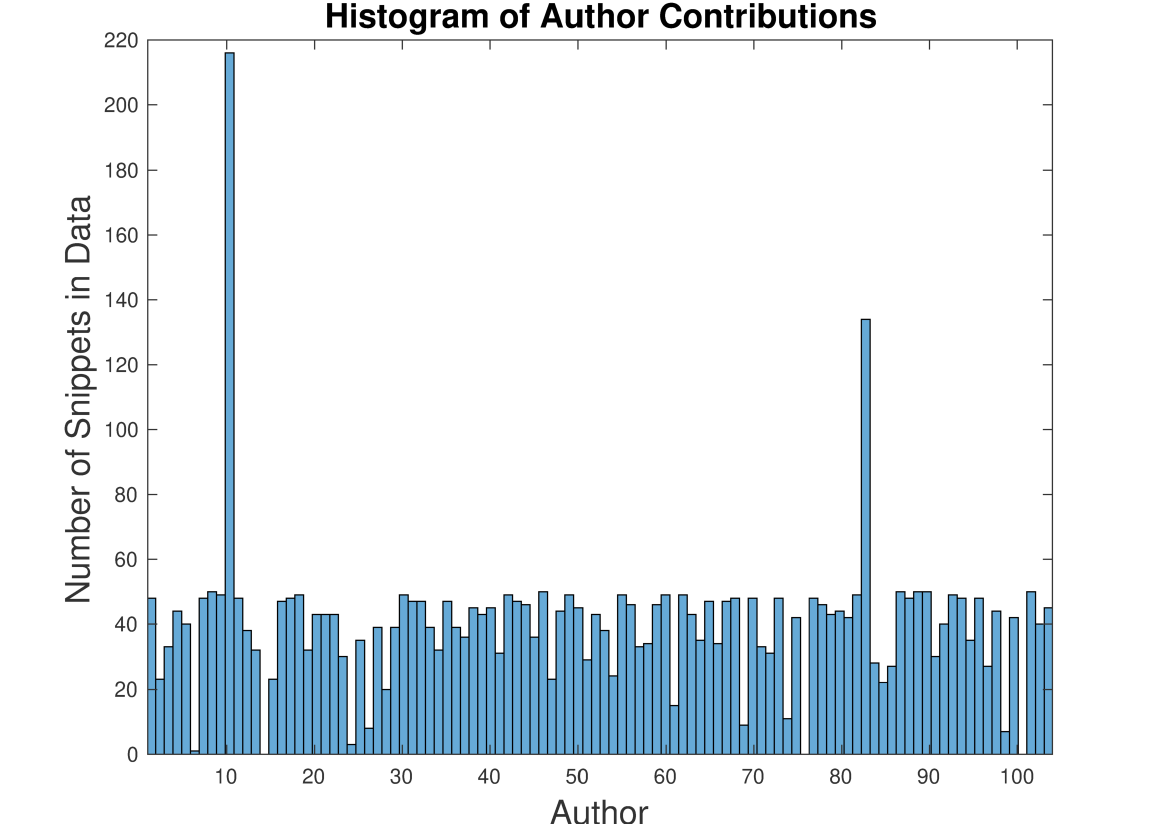}
\caption{The number of test samples per author varies from 0 to nearly 220.}
\label{fig:authData}
\end{figure}

We then removed all samples which occurred multiple times.   Our overall dataset included 104 programmers, each with at least 150 samples. For each programmer, we randomly selected files until we had 100 samples as training data, discarding left over samples from the same files.  The remaining samples, which could be fewer than 50 for authors with few files, are our ``unknown'' samples.  We repeat this split nine times and average our results across these splits.  Figure \ref{fig:authData} shows the number of samples per author in the ``unknown'' set, ranging from 0 for some authors to over 200 for others.  In total, we had "unknown" samples from 101 of our 104 authors.

While we acknowledge that the ground truth for git blame is weaker than for commits, we chose to use git blame for four primary reasons.  First, commits can include deletions and syntactically invalid constructs, both of which introduce complications in data extraction.  Second, gathering data from git blame is faster when attempting to determine authorship of current parts of chosen files, while collecting commits would be faster for attempting to determine the authorship of an individual account.  Third, we believe the results of git blame are closer to the results we would achieve if we were to use some technique to segment code which is not version controlled on a publicly accessible repository (for example, identifying differences in successive versions of source code). Fourth, we believe that in general it is more likely to want to attribute the current version of code, rather than some past version.

\subsection{Features}
\label{features}

In this work we use a feature set derived from the work of Caliskan-Islam et al.~\cite{caliskan2015anonymizing}. Our primary features come from the AST and include nodes and node bigrams \cite{aho1986compilers}. The AST, is a tree representation of source code with nodes representing syntactic constructs in the code, and we used the fuzzy parser joern to extract them \cite{YamGolArpRie14}. This parsing allows generating the AST without a complete build and without the complete build environment. Thus, it allows us to extract the AST even for our partial code samples.  We also include word unigrams, API symbols, and keywords. Our feature set includes both raw and TFIDF versions of many of our features. TFIDF, or term frequency-inverse document frequency, is a measure combining the raw frequency with a measure of how many authors use the feature. Due to the sparsity of the feature set, we do not use information gain to prune the set, and instead keep the entire feature set, minus any features which are constant across all samples \cite{quinlan1986induction}. Information gain is an entropy based measure of the usefulness of a feature for splitting data into classes by itself, and because of the sparsity of our feature vectors is zero for most features. We have a total of 451,369 features, of which 82,272 are trivially removed as zero-valued across all samples leaving 369,097 features, of which an average of 365,690 are zero-valued for any given sample.

\subsection{Single Sample Attribution}
\label{ssa}

For this case, we set aside the version control assumption and assume no external information about authorship. 
As a use case, we can imagine the case where the sample is git blamed to an anonymous user rather than to an account, or when a cautious individual is creating a new account for every commit. Therefore, we can only classify at the level of the individual sample. For this, we perform cross-validation with random forests, as in the work of Caliskan-Islam et al.~\cite{caliskan2015anonymizing}. Random forests are ensemble multi-class classifiers which combine multiple decision trees which vote on which class to assign to an instance, aimed at reducing variance while perfoming regression or classification \cite{breiman2001random,friedman2001elements}. Trees for random forests are grown to be independent and identically distributed, and the bias of each tree is equal to the bias of the forest.  By randomly selecting the inputs of each tree, the intent is to reduce correlation between trees while keeping the variance of the forest small. 

Most of our experiments use 500 random trees (200 more than used by Caliskan-Islam et al.) and $\log_{2}(M)+1$ features, where $M$ is the total number of features in the dataset. Our parameterization was based on the parameters used by Caliskan-Islam et al., expanding the number of trees to help compensate for the more difficult problem \cite{caliskan2015anonymizing}.  However, for our largest open world experiments we used only 50 random trees with 50 features each in order to show results under the circumstances in which resources (either time or computational) are limited. This serves as a baseline for our work.  Ideally, we would then continue as in Section \ref{msa} for multiple sample attribution. However, in the event that we cannot, we apply a technique to help analysts better interpret the results described in Section \ref{cc}. 

\subsection{Multiple Sample Attribution}
\label{msa}

If the version control assumption holds, such as if we observe that a group of samples belongs to a single account, we can leverage all of the samples in the group of samples to identify their collective author.

Our method requires performing the same classification as for single samples but then aggregating results for the samples which we have identified as belonging to an account.  
We aggregate the probability distribution output of the classifier rather than the predicted classes to allow more confident attributions to outweigh less confident attributions, and then take as the prediction for the aggregated samples the class with the highest averaged probability.

We also attempted a method involving merging individual samples into larger ones, but while this performed better than attributing single samples it was outperformed by the probability averaging technique.

\subsection{Calibration Curves}
\label{cc}

Because we suspect that some samples will remain which are difficult if not impossible to classify, we want to have a reliable way to separate reliable from unreliable attributions.  Towards this, we build \emph{calibration curves}, which display attribution accuracy on the y-axis and \emph{classification confidence} on the x-axis.  Classification confidence is the classifier's output probability for the selected class.  For random forests, this refers to the percentage of trees which vote for the given class.

For our analysis, we bin the samples based on the highest classifier probability output in increments of 10\%, and report the accuracy for samples in each interval.  If more granularity is needed for a particular application, we advise either decreasing the increment size or foregoing bins altogether.  While we acknowledge that the specifics of such a curve may vary for different instances of the problem and we recommend using cross-validation with known samples to prepare such a curve in order to identify a threshold for accepting a prediction based on the stakes, we expect that the overall shape of the calibration curve will remain similar for different datasets, and so ours may be used as a guide.  Through consultation with the calibration curve, analysts may determine the likelihood of accurate attribution and what level of resources needs to be dedicated to further investigation.

\subsection{The Open World}
\label{tow}

For the open world problem, in which the true author may not be in the candidate set, we use calibration curves as described in Section \ref{ssa}. We perform attribution as normal according to either single sample attribution or account attribution, and use a confidence threshold to separate samples by unknown (out of world) authors from samples correctly attributed to a suspect. This thresholding does have the drawback of grouping incorrectly attributed samples belonging to a different author in the candidate set with samples belonging to samples outside of the suspect set, but as both of these categories require further investigation we do not consider this drawback to be especially problematic.

For our initial experiment, we used 15 of the more prolific programmers with 250 samples available for training data as our suspect set, with a small set of samples from authors outside that set.

For our main experiments, we divided our 104 programmer dataset into four disjoint subsets of 26 programmers. For each of four rounds, we took one of those sets as our suspect set $S$, and the remaining 78 programmers as the set of unknown authors $U$. We performed the same analysis on the samples belonging to the authors in $S$ as before, adding all of the documents from $U$ to each evaluation set.  We then binned the samples as in our calibration curves, with each bin maintaining counts of correct attributions, incorrect attributions of samples belonging to authors in $S$, and samples belonging to authors in $U$, which we refer to as being ``out of world''. We chose this set-up to simulate the case in which an analyst only cares if one of a small number of people was responsible for a given account or code sample.
We note that devising our experiments in this way allowed us to heavily bias our evaluation set in favor of samples by programmers outside of our suspect set, analyzing 370,529 out of world samples and 29,904 ``in world'' samples between all rounds, for a total of 400,443 samples of which 92.5\% are out of world.  

We then evaluated thresholds at the lower bound of each bin in terms of precision and recall, but rather than calculate the precision and recall of the classifier itself we computed precision and recall with respect to the three classification counts maintained by the bins, using the threshold as the selector. Precision is a measure of the percentage of selected instances which belong to the desired category while recall is a measure of the percentage of instances belonging to the desired category which were selected.  We calculated precision and recall according to the following three criteria: correct classifications above the threshold, out of world samples below the threshold, and samples which are either out of world or classified incorrectly below the threshold.

For real world applications, we suggest using a similar approach, evaluating on a separate labeled dataset containing some samples from the suspects as well as samples from numerous other authors, prior to introducing the actual samples of interest.  We then suggest setting a threshold based on these values and the needs of the application, and then determine where the samples to be attributed fall in order to determine whether to conclude the attribution is correct or incorrect, with the goal that out of world samples will be determined as incorrectly attributed.  
For the purposes of this paper, however, we calculate the \emph{F1 score} for each of the three criteria and use this to predict what may be useful thresholds, and to evaluate the power of the technique.  \emph{F1 scores} are calculated according to the formula $2*(precision*recall)/(precision+recall)$, and are a harmonic average of the two values.

\section{Results}
\label{sec:results}

From the 104 programmer dataset, our baseline single sample attribution accuracy was 48.8\%, or 61.4\% when averaging over authors, compared to 5.5\% accuracy if we attribute every sample to the author with the most samples. Averaging over authors is also known as \emph{balanced accuracy} and compensates for unbalanced testing classes by computing accuracy for each author and then averaging those accuracies \cite{brodersen2010balanced}.  We note that while this is much lower than the accuracy reported by Caliskan-Islam et al.~\cite{caliskan2015anonymizing}, the data itself is very different. That work attributes whole source code files written privately with an average of 70 lines of code per file, while our work attributes pieces of files written publicly and collaboratively with an average of 4.9 lines of code per file. Intuitively, it is reasonable to believe that our dataset contains samples which are much harder to classify. The difference in accuracy based on how we average means that overall we have more samples belonging to more difficult to attribute authors. We will use both measures going forward, to illustrate both a lower and higher estimate of accuracy. While not part of our main results, we show basic results of relaxed attribution in this environment for those interested in Appendix \ref{sec:relax}.

\subsection{Multiple Sample Attribution}

\begin{figure}[htb]
\centering
\includegraphics[width=\columnwidth]{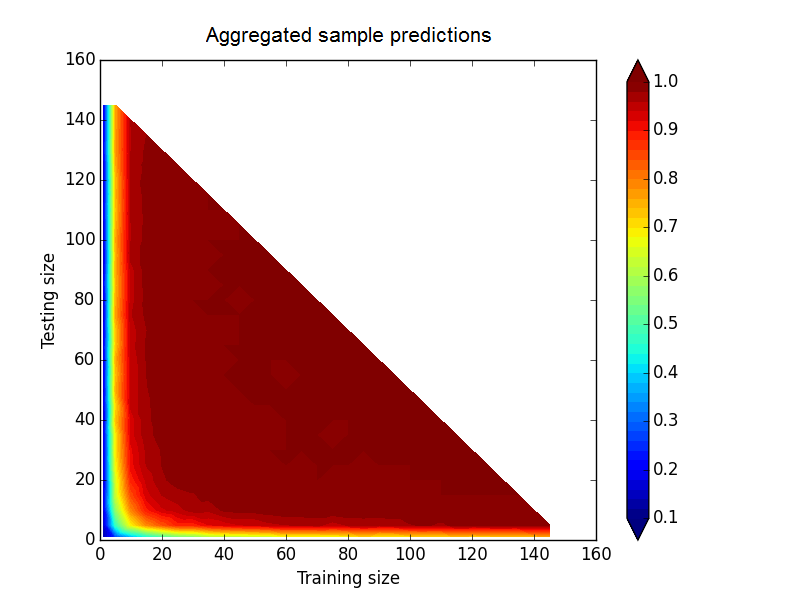}
\caption{Accuracy changes as we vary the number of training instances and the number of aggregated classification samples, with the units on both axes representing the number of samples.}
\label{fig:15progAgg}
\end{figure}

Before starting experiments with the full dataset, we performed preliminary experiments with a small subset to determine how much data would be necessary for our aggregation method. Figure \ref{fig:15progAgg} shows the results of varying the number of training samples and aggregated result classification samples on a small subset of our data with only 15 programmers using standard cross-validation, which we note can create the situation in which we train and test on code segments originating in the same file, inflating our results and permitting over-fitting. However, this can simulate the case in which we know the identity of the author of most lines in a file, but are attempting to attribute a small number of lines of code with unknown authorship. 
The inflation of accuracy due to cross-validation may change the specific values, in particular requiring more training data to better express a general style, but our main experiments show that the overall relationship between the number of samples of both training and testing data and accuracy remains.

\begin{figure}[htb]
\centering
\includegraphics[width=\columnwidth]{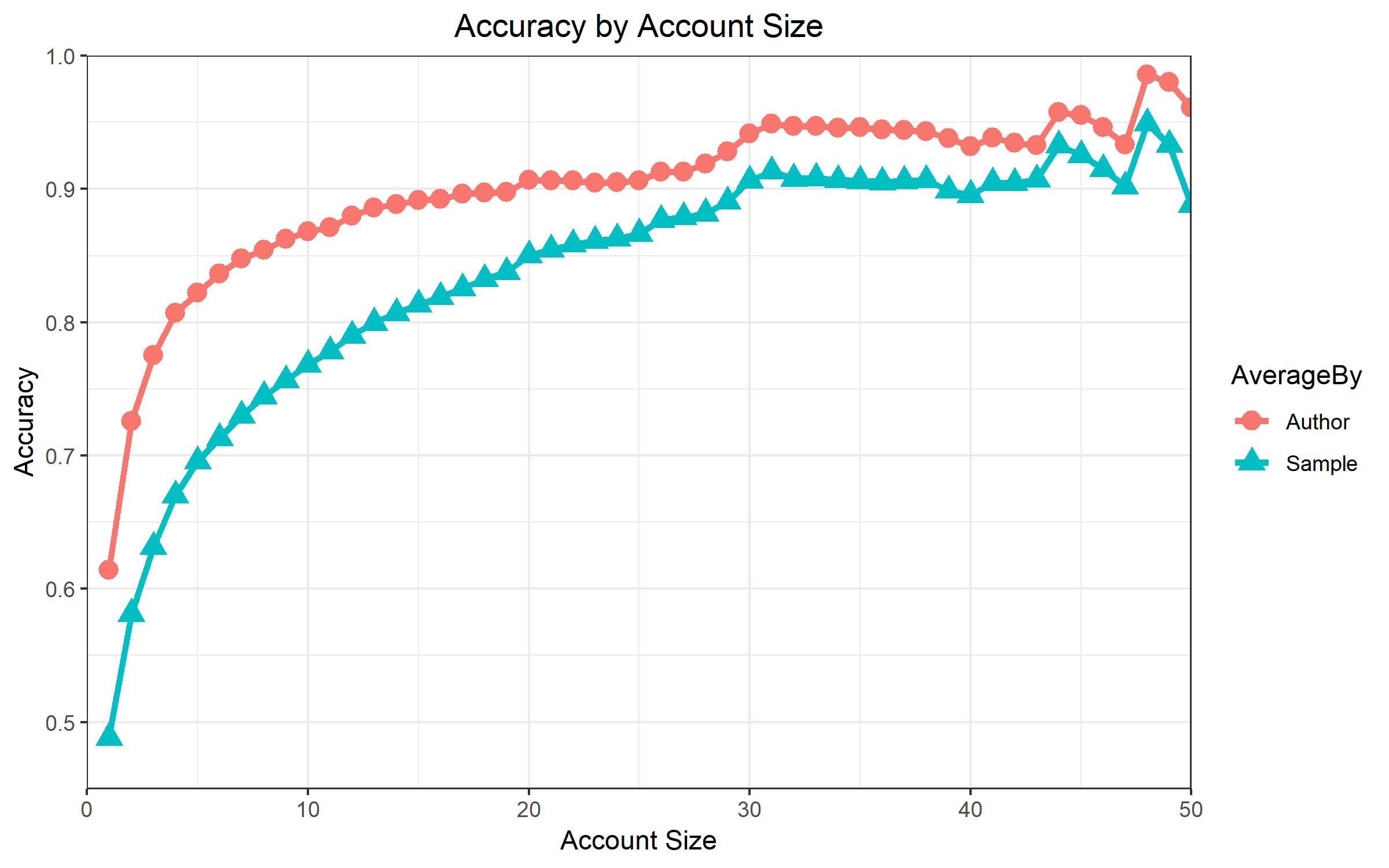}
\caption{Accuracy changes (mostly increasing) as we increase from single sample attribution to aggregating 50 samples.}
\label{fig:106authorsstrong}
\end{figure}

Figure \ref{fig:106authorsstrong} shows the results for varying numbers of aggregated results from our full 104 programmer dataset.  
We note that as we increase account size, or the number of linked test samples for multiple sample attribution, we drop from the evaluation set authors with insufficient samples, which partially explains the lack of smoothness in the curve, especially as we approach 50 samples per account, as many authors have fewer than 50 remaining test samples as shown in Figure \ref{fig:authData}. 
We note that in many cases, a single collaborative code file will contain multiple samples for most programmers, and 15 samples is not unreasonable for larger projects or for programmers who contribute to multiple projects.

\subsection{Calibration Curves}

\begin{figure}[htb]
\centering
\includegraphics[width=\columnwidth]{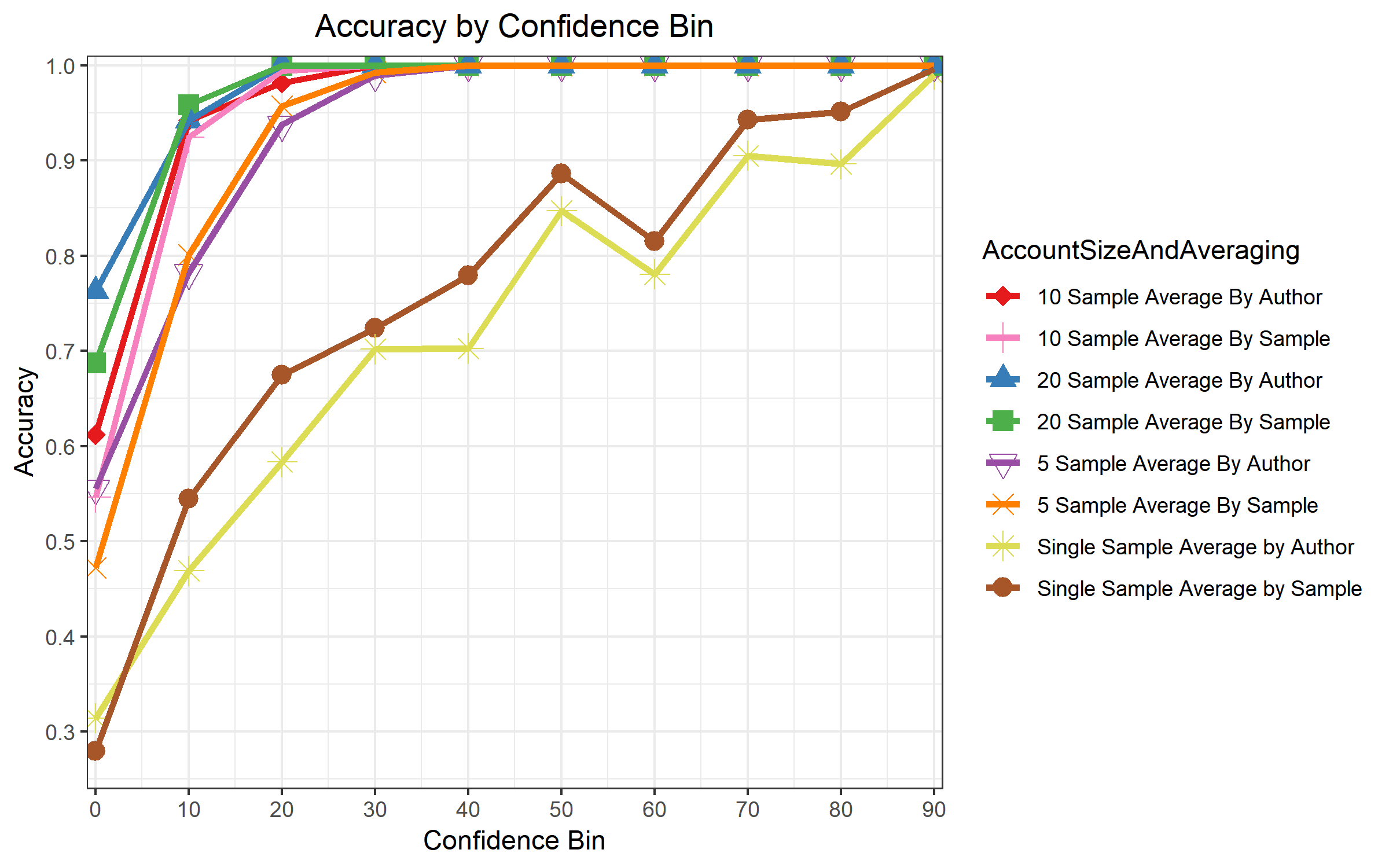}
\caption{Accuracy mostly increases with the confidence level of the classifier for sample averaged and author averaged accuracy for accounts of sizes 1, 5, 10, and 20 samples.  The x-axis shows the probability for the predicted class as output by the classifier, divided into bins of 10\%, starting from the displayed value.}
\label{fig:calibration106}
\end{figure}

Figure \ref{fig:calibration106} shows the calibration curve constructed from our experiments. For ease of reading, we only show single sample and accounts of sizes 5, 10, and 20 samples. The calibration curve shows that our classifier is conservative: the predicted probability is lower than the actual accuracy obtained (a known feature of random forests). Even when the classifier confidence is less than 10\% we still outperform random chance. For account attribution, we notice that even low confidence attributions are highly accurate.  We note that our experiments included even larger accounts, but at large accounts we did not have attributions within all confidence intervals - typically we only had attributions at less than 40\% confidence and accuracy well over 90\%, and in some cases some attributions at over 90\% confidence, with nothing in between.

\begin{figure}[htb]
\centering
\includegraphics[width=\columnwidth]{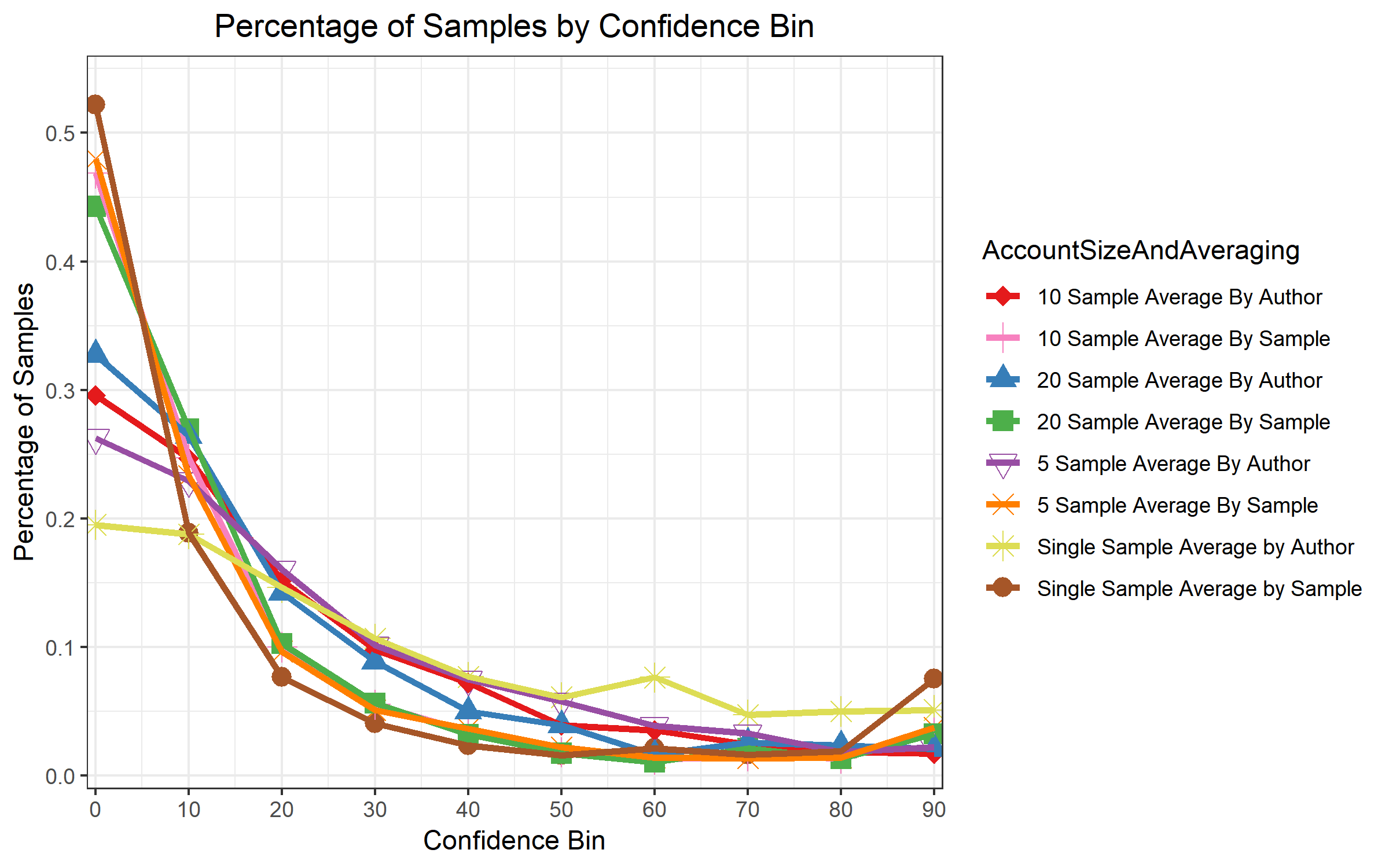}
\caption{The percentage of the samples in each of the 10\% confidence intervals for the calibration curves in Figure \ref{fig:calibration106}, starting from the displayed value, mostly decreases as confidence increases.}
\label{fig:calibSample}
\end{figure}

We also counted the number of samples which fell into each confidence interval for the data set.  Figure \ref{fig:calibSample} shows the percentages of samples which fall in each interval.  For the 104 programmer dataset, most of the samples fall in the lower confidence intervals, which explains why our accuracy is low compared to the median bin accuracy. We also notice that the sample confidence distribution is more heavily skewed towards low confidence than the author confidence distribution, again suggesting that our dataset contains more samples from more difficult authors than easier authors.

Our calibration analysis suggests that our low overall accuracy compared to previous results on full source code files from Google Code Jam is likely in part a result of our dataset containing data which is difficult to classify.  

The calibration curve also suggests another potential reason that our classification result aggregation method works so well.  Because our classifier is so conservative, our mis-attributions tend to have very even spreads of low probabilities compared to our correct classifications, which means that they do not easily outweigh the probabilities of correctly classified instances.  This leads to the whole group being correctly classified.

\subsection{The Open World}
\label{open-world}

\begin{figure}[htb]
\centering
\includegraphics[width=\columnwidth]{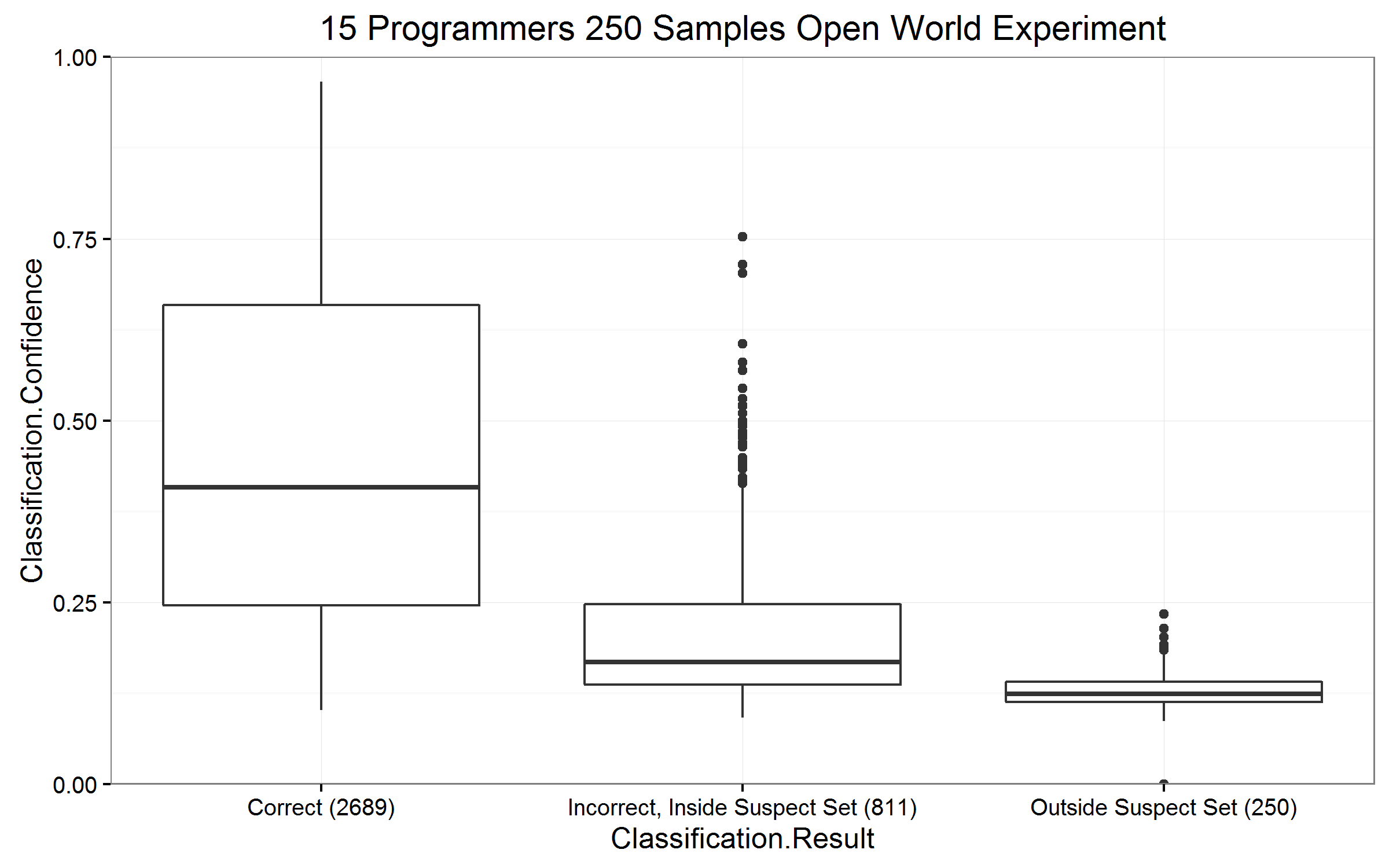}
\caption{Classifier confidence distributions differ between correct classifications, mis-classifications of samples belonging to suspect authors, and out of world samples using a suspect set of 15 programmers with 250 samples each with an additional 250 out of world samples. 
The x-axis labels also include the number of samples which belong to each group.}
\label{fig:confidence}
\end{figure}

 Figure \ref{fig:confidence} shows the results of our initial open world experiment. In this experiment, we used the 15 programmer dataset as our suspects and trained a model on these programmers. We then tested this model on a few samples from the remaining programmers in our larger dataset to create open world conditions.  Our results suggest that the calibration curve method is a viable way to address the open world problem.  We notice that for our dataset, samples which do not belong to the suspect set usually have classification confidence below 20\% and the highest such confidence is 23\%, while most incorrect classifications occur with confidence below 40\%.  

We also notice that the percentiles tend to match up with each other between classification results, however this is likely a quirk of the specific dataset and experiments.   
Our results suggest that mis-attributions due to the open world scenario are similar to general mis-attributions with respect to classification confidence and that an open world can similarly be handled by discarding low confidence predictions, which we would likely already discard or handle skeptically based on our calibration curve (see Figure \ref{fig:calibration106}). 

For the remaining open world experiments, we used the full 104 programmer dataset, with 26 programmers in the suspect set and 78 in the out of world set for each run. We used 50 tree random forests with a limit of 50 features per tree, as a weaker classifier which performs faster and uses fewer computational resources.

\begin{figure}[htb]
\centering
\includegraphics[width=\columnwidth]{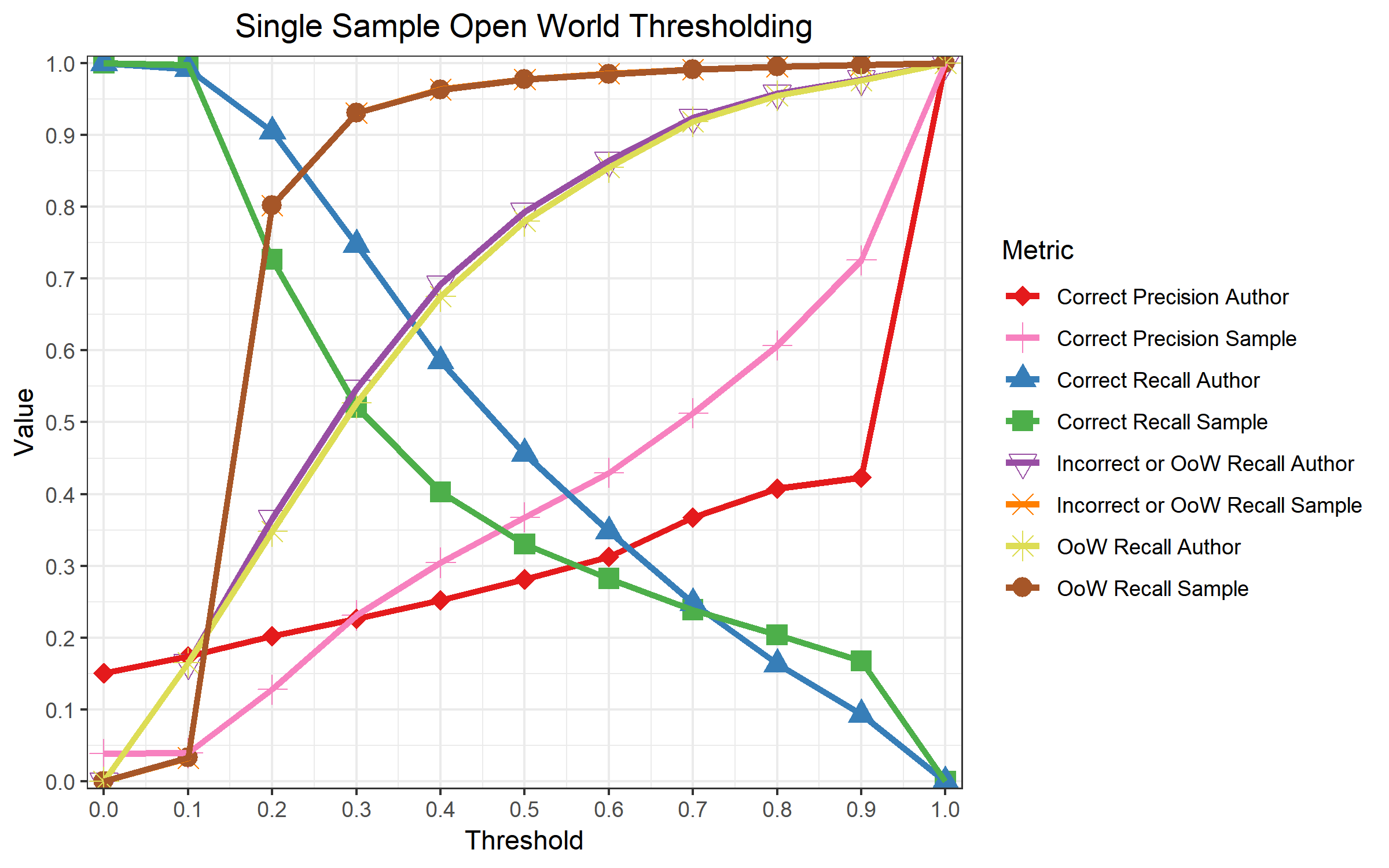}
\caption{Precision and recall change for our open world experiments for single sample attribution as we increase the acceptance/rejection threshold.}
\label{fig:ow10}
\end{figure}

\begin{figure}[htb]
\centering
\includegraphics[width=\columnwidth]{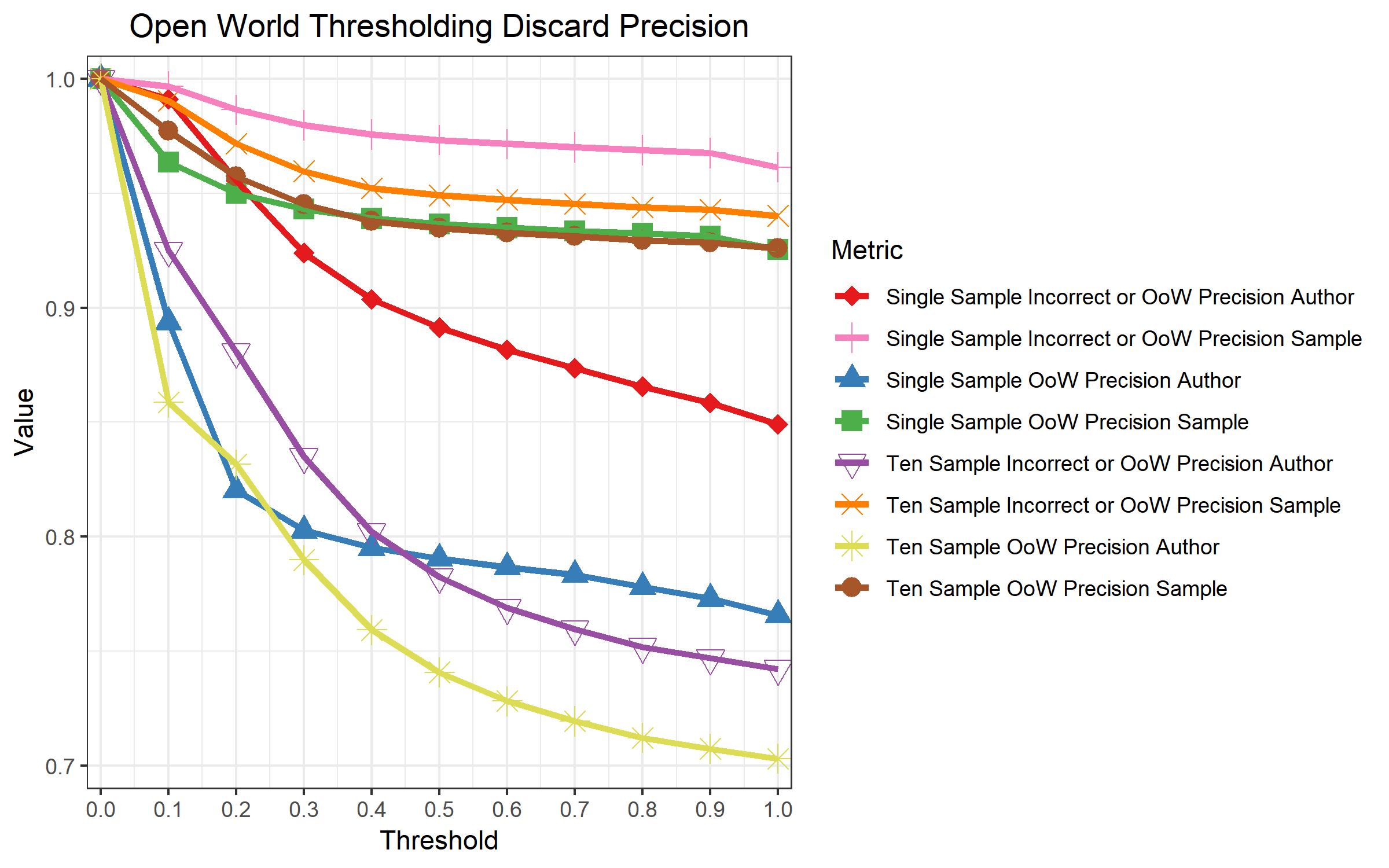}
\caption{Precision for discarding incorrect attributions and out of world samples in both single attribution and 10-sample account attribution decreases as we increase the rejection threshold.}
\label{fig:owdp}
\end{figure}

Figure \ref{fig:ow10} shows the results of our open world experiments using 26 suspect programmers and 78 unknown programmers for the more difficult single sample attribution case with respect to precision and recall of identifying correct or incorrect attributions using a threshold.  We note that the overall accuracy, ignoring the out of world samples, is 51.7\% by sample or 64.5\% by author for 26 suspects with the weak classifier.  We also leave precision for identifying bad attributions for a separate graph combined with precision for identifying bad attributions with account attribution, as these values were consistently high, which can be found in Figure \ref{fig:owdp}.  As an example for reading the figures, at a threshold of 0.4 confidence and averaging by sample (discarding all attributions with confidence under 0.4 and keeping all attributions with confidence of at least 0.4), recall shows that we keep 40.3\% of correct attributions and discard 96.3\% of incorrect attributions and out of world samples while precision shows that of the samples we keep 30.5\% are correct attributions and 97.6\% of samples discarded are incorrectly attributed (93.9\% are out of world).

We consider selecting a threshold to be application specific because both the importance and values of the precision and recall measures may vary. With respect to discarding bad attributions, discarding all attributions always had the highest F1 score (.984 for identifying incorrect or out of world samples averaged by sample, .867 for identifying purely out of world samples averaged by author, and 1 otherwise), likely due in part to the heavy weighting of the evaluated dataset in favor of out of world samples. However, with respect to keeping correct attributions, the best F1 score occured at a threshold of either 40\% or 50\%, depending on averaging.  If averaging by sample, the best threshold was 50\% with an F1 score of .348.  If averaging by author, the best threshold was 40\% with a F1 score of .352. 
As we increase the threshold the precision for correct attributions rises because as the confidence increases it becomes much more likely that the attribution was correct. However, the recall for correct attributions falls because we have many samples which are attributed with low confidence, which include many correct attributions once we reach thresholds of 20\% and higher. For the out of world and incorrect samples, we notice that precision is consistently high but falls slowly. When combined with the fact that the recall rises sharply early before leveling off, this suggests that we quickly identify the majority of the out of world and incorrect samples while discarding relatively few correctly attributed samples, and so once we go beyond a threshold of about 40\% or 50\%  confidence we are mostly losing correct attributions and not identifying as many out of world or incorrect attributions. Taken together, this reinforces what we noticed previously in Figure \ref{fig:confidence}: correct attributions have a different, although overlapping, confidence distribution from incorrect attributions and out of world samples.

\begin{figure}[htb]
\centering
\includegraphics[width=\columnwidth]{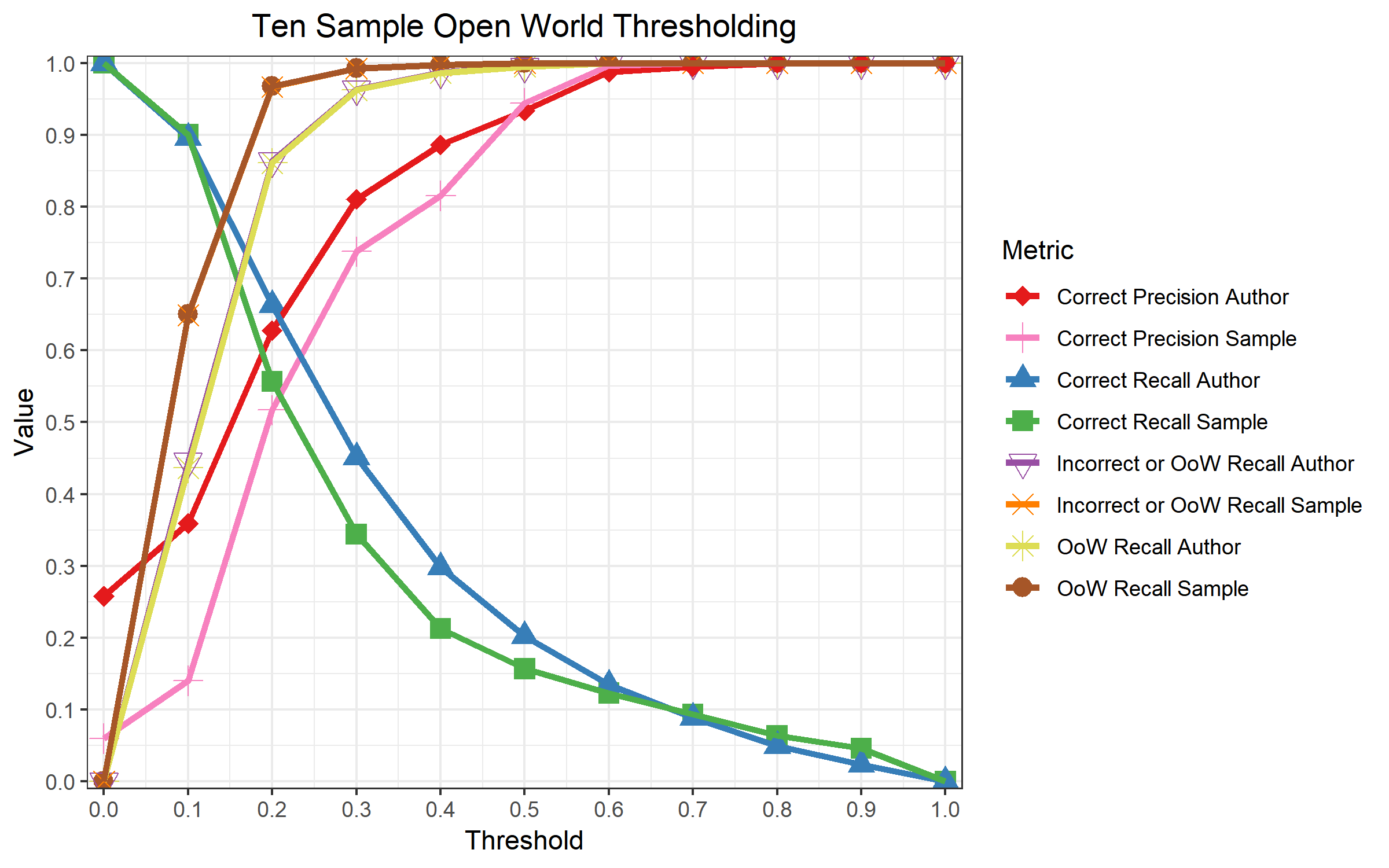}
\caption{Precision and recall change for our open world experiments for account attribution for collections of 10 samples as we increase the acceptance/rejection threshold.}
\label{fig:7ow10}
\end{figure}

Figure \ref{fig:7ow10} shows the results of our open world experiments with 26 suspect programmers and 78 unknown programmers for the multiple sample attribution case for collections, or accounts, of 10 samples.  We chose 10 samples because at that level we still have over 400,000 observations, and note that different account sizes will have different values but similar trends, with the level of improvement over the single sample attribution case generally corresponding to the size of the account.  We note that the overall accuracy, ignoring the out of world samples, is 80.8\%.  The optimal F1 score for correctly attributed instances above the threshold is 20\%, with F1 score of .536 averaging by sample or .645 averaging by author. The optimal F1 score for discarding samples which are either out of world or otherwise incorrectly attributed is 30\%. If we are only interested in out of world samples, the F1 score is .968 if we average by sample or .868 if we average by author.  If we are interested in all misattributions, the F1 score is .976 if we average by sample or .895 if we average by author. 
We also experimented with collections of 5 samples, with the same trends but slightly lower accuracy and F1 scores.

\begin{figure}[htb]
\centering
\includegraphics[width=\columnwidth]{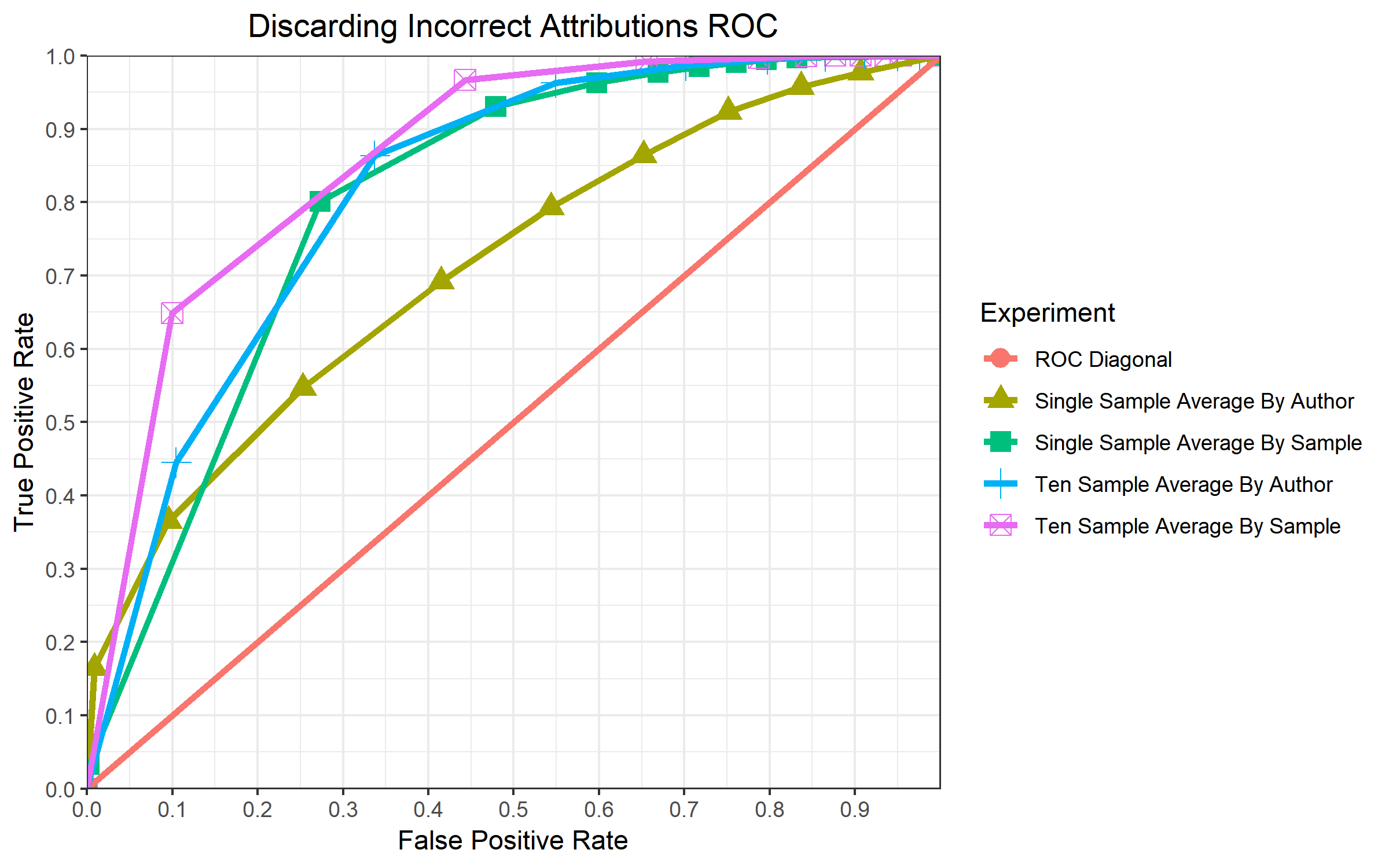}
\caption{This ROC curve compares false and true positive rates for identifying false attributions, either due to out of world samples or incorrect classifications, by setting a threshold.}
\label{fig:roc}
\end{figure}

Figure \ref{fig:roc} shows a receiver operating characteristics (ROC) curve for the task of identifying false attributions using a threshold. ROC curves plot true positive rate on the y-axis and false positive rate on the x-axis. This analysis can help evaluate acceptable trade-offs, which can then assist in choosing a threshold.  For this curve a true positive is a correctly identified false attribution, while a false positive is a correct attribution mistakenly determined to be false. For example, focusing on the 10 sample account attribution averaged by sample, we can see that by discarding only 10\% of the correct attributions we can discard 64.9\% of the incorrect attributions and by discarding 44.3\% of the correct attributions we can discard 96.7\% of incorrect attributions.

From these experiments, we can conclude several things.  First, while there are cases in which out of world samples may be mistaken as belonging to one of the suspect programmers by using this technique, these cases are relatively rare.  As we increased the threshold we started with a few dramatic cuts to the percentage of out of world samples above the the threshold, correctly identifying over 90\% of such samples in only a few increments for even the harder problem of single sample attribution and then continuing to identify about 97\% and then over 99\% as we continue to increment.

In all of these experiments, while raising the threshold causes us to doubt correct classifications, it allows us to correctly identify incorrect classifications and out of world samples more quickly than we reject correct attributions. While we observe that our technique cannot completely separate correctly classified samples from either incorrectly attributed samples or out of world samples, we can use it to easily find trade-offs that allow us to identify most out of world samples and to trust our remaining attributions to a high degree.  It is also notable that our results suggest that the reason it is hard to separate correctly attributable samples from out of world samples is not because it is hard to identify most of the out of world samples but because some samples are harder to attribute than others.  We can also observe that easier problems allow for lower confidence thresholds and fewer correct attributions misidentified as incorrect attributions, with the threshold for 10 sample account attribution requiring classifier confidence of only about half that needed for single sample attribution with similar results. 

We note again that our open world experiments were done with a weaker version of the classifier, with fewer trees of limited depth in the random forest, and with 92.5\% of evaluation samples by programmers outside of the suspect set. We would expect a stronger version of the classifier to reach the levels of success found in our experiments at lower thresholds.

\section{Discussion and Analysis}
\label{sec:Analysis}

While the above results show that we are able to attribute source code in the wild with a high degree of accuracy, they also suggest that a major difficulty in attributing source code samples is the high within-author variance of such samples: short segments of code from a single author may be distributed across a range of tasks and thus contain limited and in many cases ambiguous stylometric information.  

\subsection{Attributibility Analysis}

We performed a preliminary analysis on the simple cross-validation results, and we noticed some characteristics shared by many, although not all, of the mis-attributions.  Many of the misclassified samples were trivial, and contained only very basic programming structures.  The majority of the misclassified instances had only a few abstract syntax tree nodes per line of code, with many of the longer samples averaging less than one node per line of code. 57.4\% of the misclassified samples had only 1 line of code, and 43.4\% of the of the samples with only 1 line of code were misclassified. The average length of misclassified samples was 3.7 lines of code, while correctly classified samples were 5.7 lines of code long on average.
Thus many of our misclassified samples have only a few abstract syntax tree nodes and most of the information comes from the specific word unigrams which make up the code.  As noted in the work of Caliskan-Islam et al.\ \cite{caliskan2015anonymizing}, word unigrams provide less information than AST nodes, and intuitively word unigrams are easier to obfuscate and are often standardized in collaborative code projects.  Additionally, word unigrams may contain information such as comments, licensing, and hard-coded strings.  Therefore, it is to be expected that samples for which most of the already small amount of information comes from word unigrams rather than from AST nodes would prove more difficult to classify.

This intuition is further supported by feature analysis on the dataset, as shown in Table \ref{tab:FeatureImportance}. For this analysis, we reduced the feature space to features occurring at least 6 times in the dataset, removing the most unique features. We see that word unigrams dominate the feature importance, with AST node bigrams coming in second. A manual analysis of samples reveals that the easiest attributions were of samples including such content as licensing and highly specific functions, while the most confident misattributions included similar content from atypical authors. Similarly, the most difficult attributions included highly generic function calls. Based on this, it seems that due to the lack of available information, the classifier relies on such highly specific features, making more general attribution more difficult.

\begin{table}[]
\centering
\caption{Feature analysis}
\label{tab:FeatureImportance}

\begin{tabular}{|l|r|c|}
\hline
FeatureType & NumFeatures & FeatureImportance \tabularnewline
\hline
\hline
ASTFunctionIDCount & 1 & 0.0029\tabularnewline
\hline
ASTNodeBigramTF & 47910 & 0.1815 \tabularnewline
\hline
ASTNodeTypeAvgDepth & 16231 & 0.0733 \tabularnewline
\hline
ASTNodeTypeTF & 13538 & 0.0672 \tabularnewline
\hline
ASTNodeTypeTFIDF & 11944 & 0.0285 \tabularnewline
\hline
CFGNodeCount & 1 & 0.0027 \tabularnewline
\hline
CPPKeyword & 74 & 0.0146 \tabularnewline
\hline
FunctionIDCount & 1 & 0.0007 \tabularnewline
\hline
MaxASTLeafDepth & 1 & 0.0017 \tabularnewline
\hline
WordUnigrams & 49424 & 0.5584 \tabularnewline
\hline
\hline
Total: & 139125 & 0.9316\tabularnewline
\hline
\end{tabular}

\end{table}

\begin{figure}[tb]
\centering
\includegraphics[width=\columnwidth]{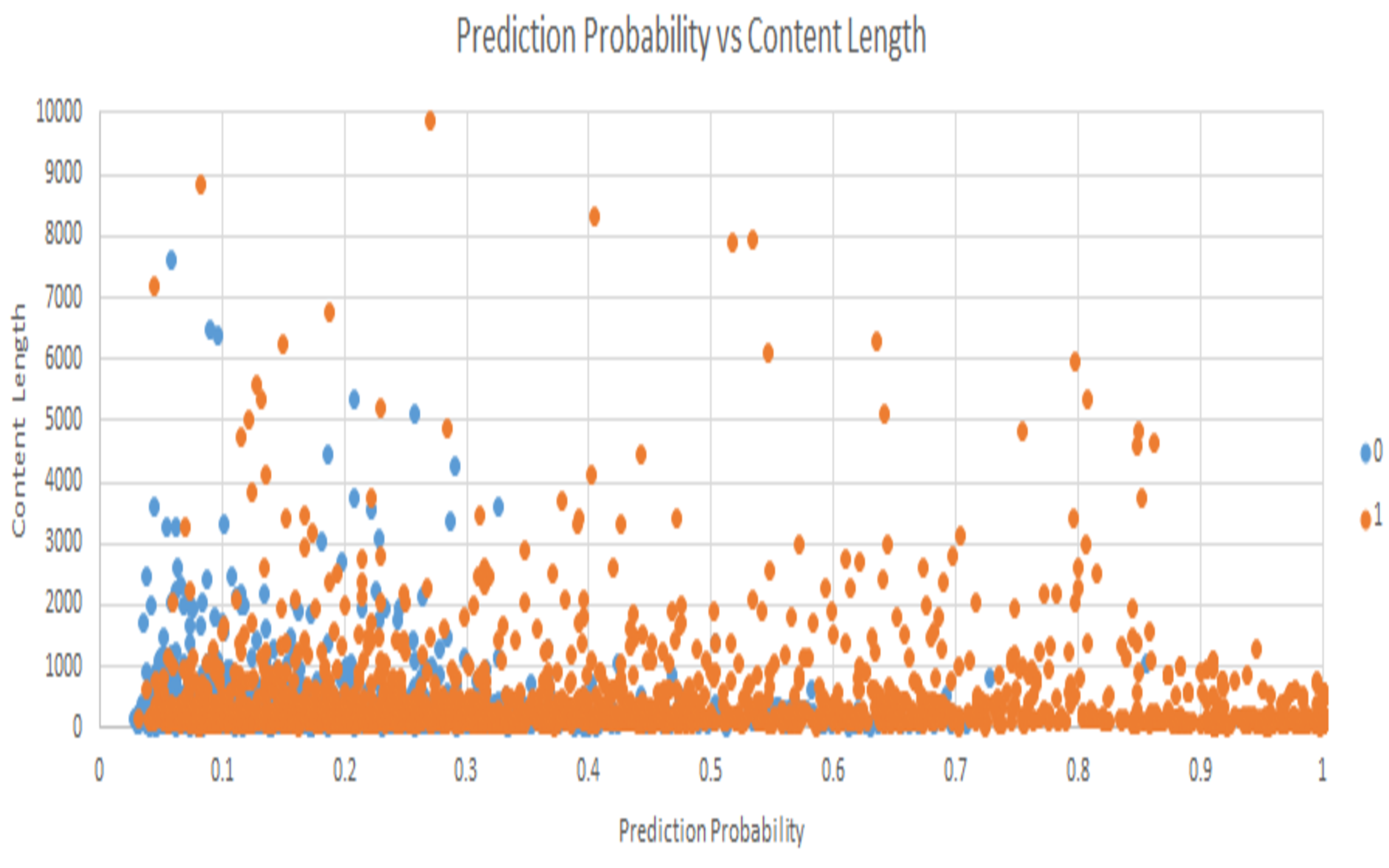}
\caption{This plot shows correct (labeled 1) versus incorrect (labeled 0) attributions by classifier confidence on the x-axis and sample length in characters on the y-axis.}
\label{fig:lvc}
\end{figure}

\begin{figure}[tb]
\centering
\includegraphics[width=\columnwidth]{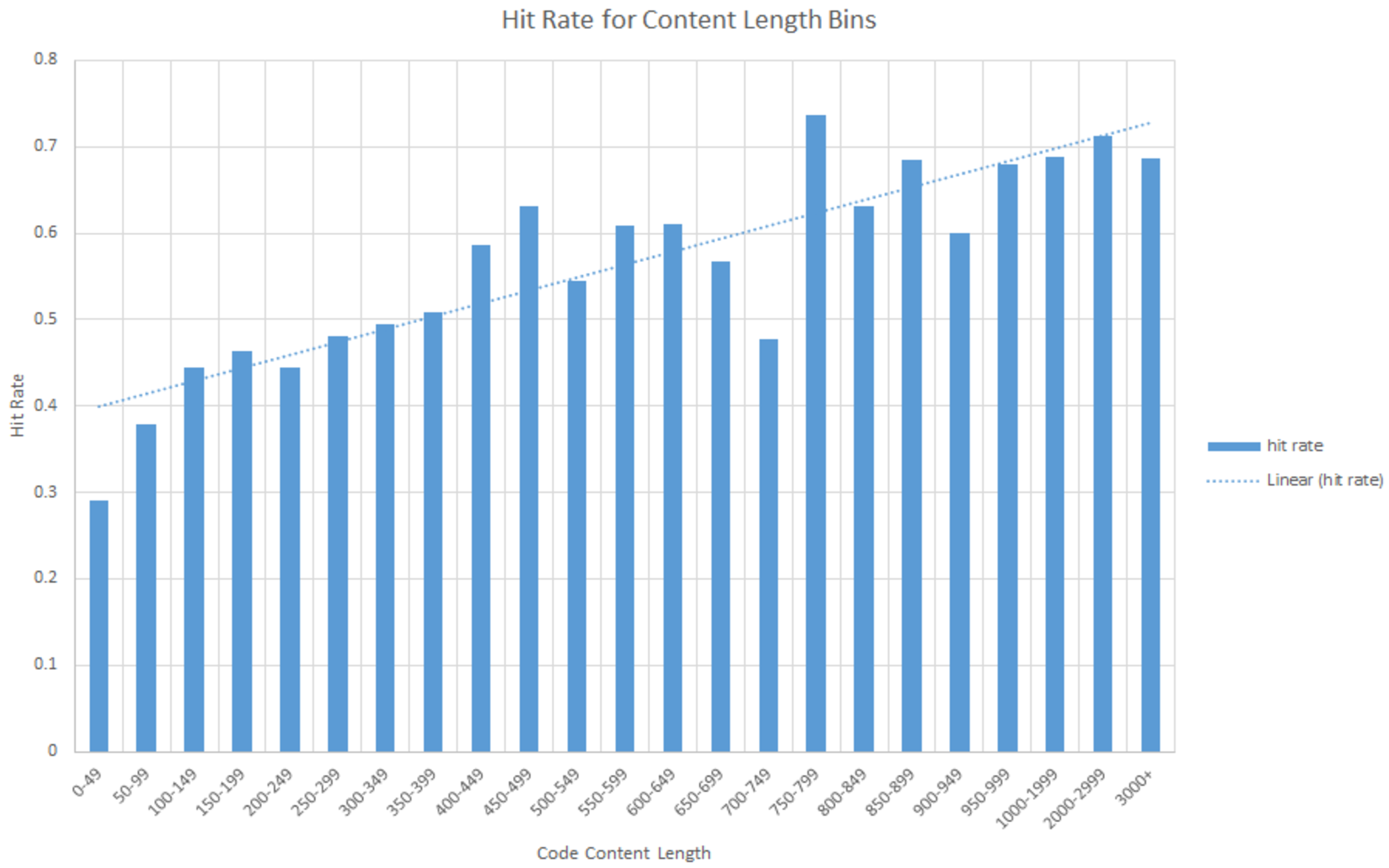}
\caption{This plot shows accuracy, or hit rate, for samples grouped by the number of characters.}
\label{fig:binhit}
\end{figure}

\begin{figure}[tb]
\centering
\includegraphics[width=\columnwidth]{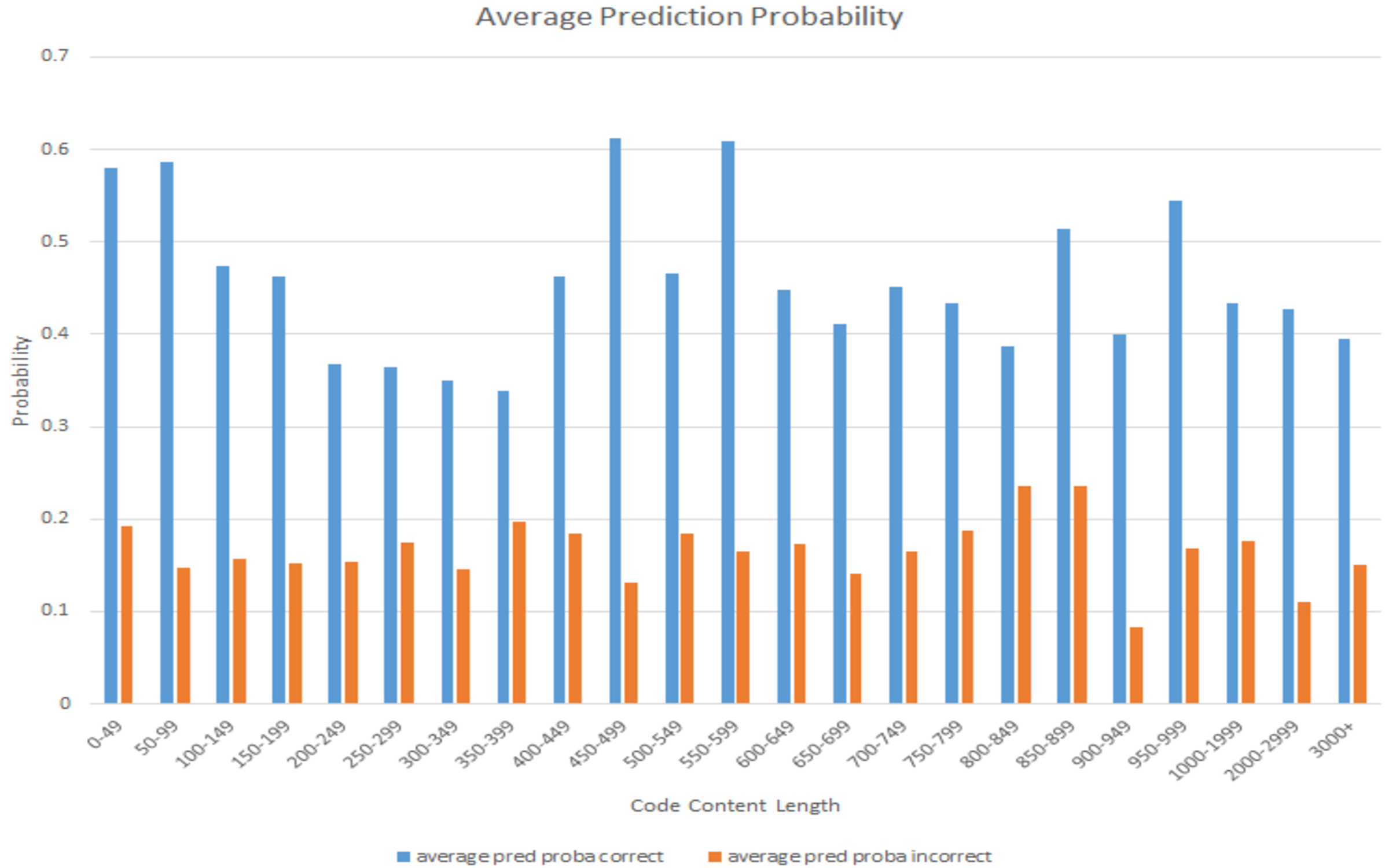}
\caption{This plot shows the average prediction probability for correct (in blue) versus incorrect (in orange) attributions grouped by the number of characters in the sample.}
\label{fig:binprob}
\end{figure}

As we have identified a lack of information per sample as a primary limiting factor, we considered whether length of samples could be an indicator of success. We measured length by the number of characters in the sample.  Figure \ref{fig:lvc} shows each evaluation sample plotted by the classifier confidence and length, with the color indicating correct or incorrect attribution, and Figure \ref{fig:binhit} shows the accuracy divided into bins by the length of the sample.  Figure \ref{fig:binprob} shows the classifier probability for individual samples grouped by the number of characters.  We note that there does not appear to be a strong relationship between sample length and classifier probability, and, while we can see a relationship between sample length and accuracy, classifier confidence remains a stronger estimator.  The observation that both sample length and classifier confidence have relationships to accuracy but lack a clear relationship to each other suggests that they may be able to be combined to create an even stronger method of differentiating correct from incorrect attributions, but this remains an extension for future work.

\begin{figure}[tb]
\centering
\includegraphics[width=\columnwidth]{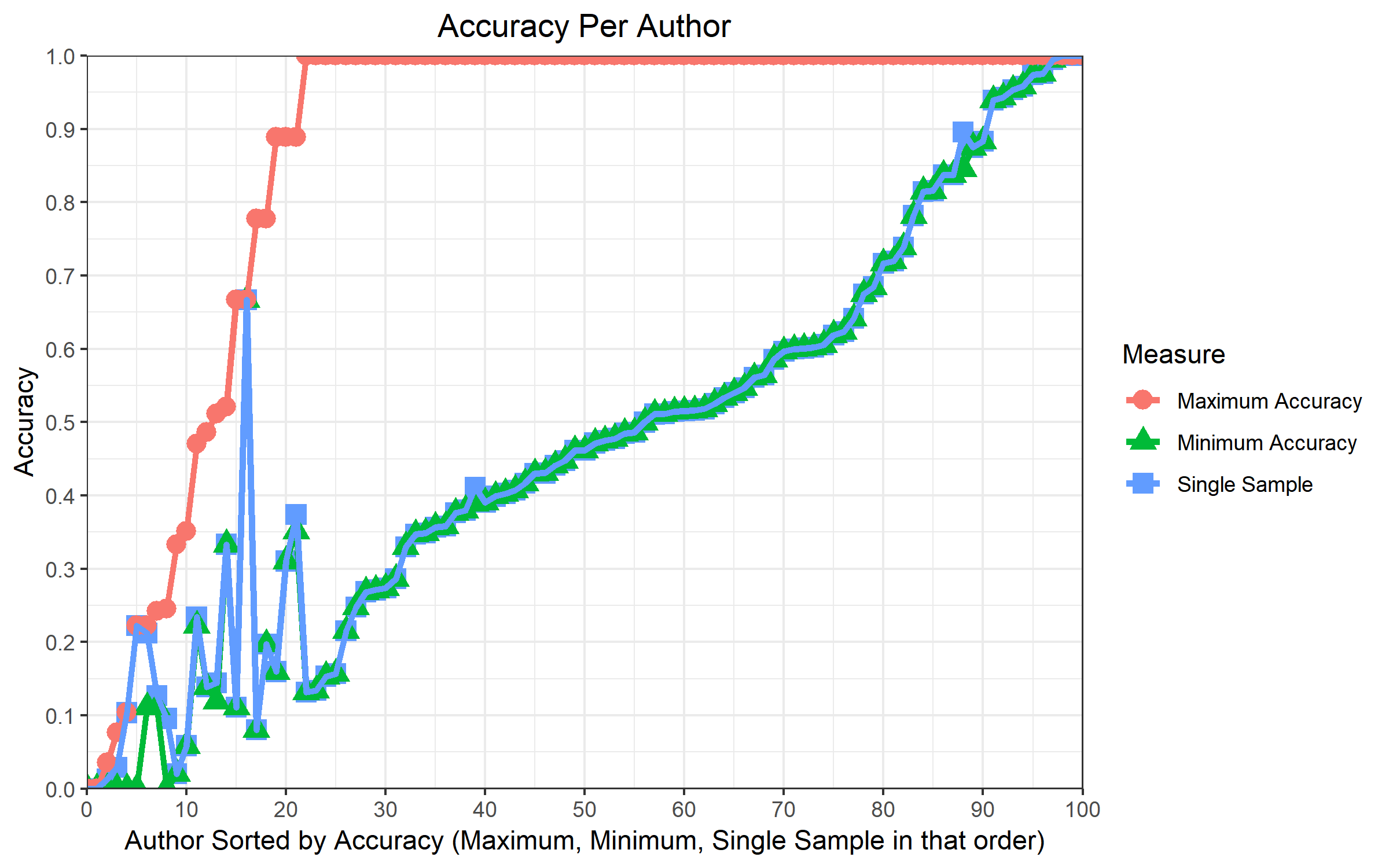}
\caption{This plot shows the minimum account accuracy (the smallest observed accuracy for an account of any size belonging to that author), maximum account accuracy (the largest observed accuracy for an account of any size belonging to that author), and single sample attribution accuracy for each author with files in the evaluation set. Authors are sorted first by maximum account accuracy, then minimum account accuracy, and then single sample accuracy.}
\label{fig:authAcc}
\end{figure}

As we have also identified that some authors are easier to attribute than others, as is made clear by the change in accuracy according to the two methods of averaging our samples, we decided to plot the accuracy per author.  Figure \ref{fig:authAcc} shows the single sample attribution accuracy, minimum account attribution, and maximum account attribution accuracy.  We note that minimum account attribution accuracy usually, but not always, occurs at account size of one (single sample attribution).  Similarly, maximum account attribution accuracy usually, but not always, occurs at the largest possible account for that author, although it may also occur with smaller accounts as well. Notably, we observe two authors we never predict correctly even with account attribution and three authors we always predict correctly even with single sample attribution.

\begin{figure}[tb]
\centering
\includegraphics[trim={1cm 7cm 1cm 7cm},clip, width=\columnwidth]{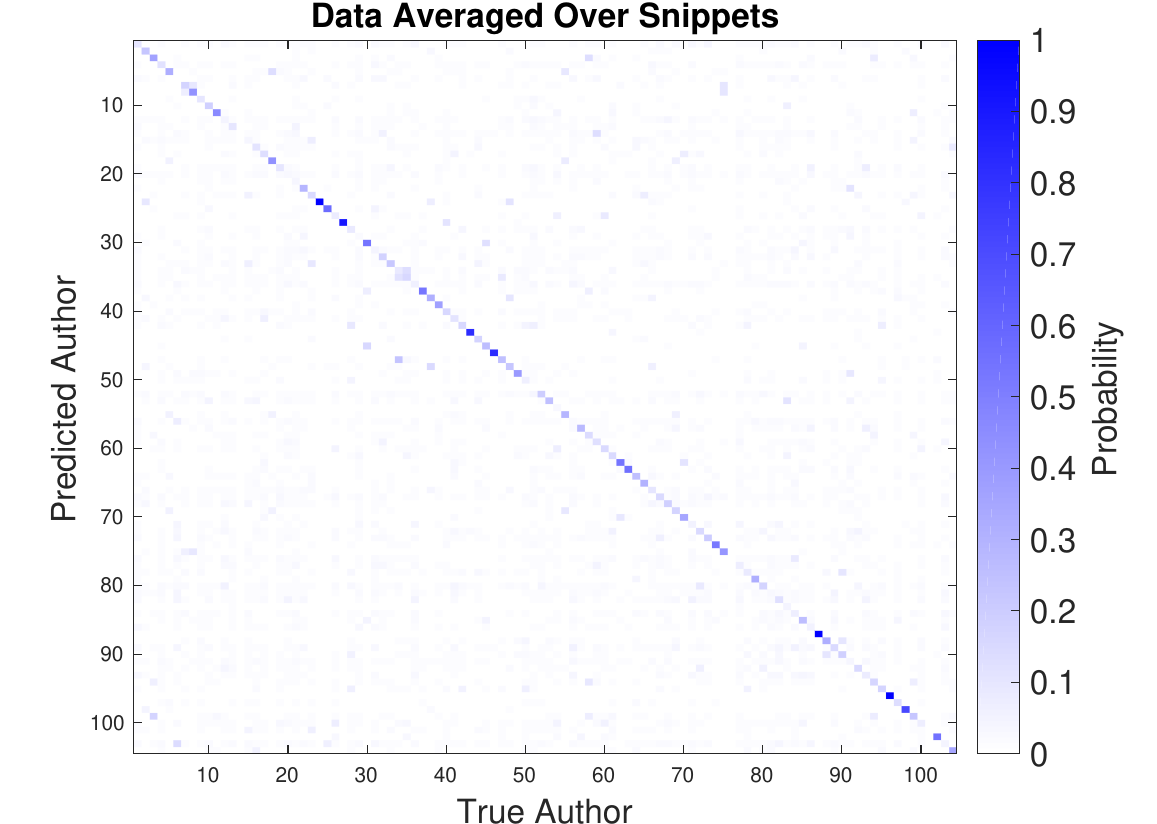}
\caption{This is a graphical representation of the confusion matrix, with the true author on the x-axis and the predicted author on the y-axis. Darker spots indicate a higher proportion of the instances belonging to the true author were assigned to the predicted author.}
\label{fig:confGraph}
\end{figure}

We also examined the confusion matrix to determine if the misclassifications are distributed more or less randomly or if they are highly concentrated. A confusion matrix shows how each sample is classified compared to the true class. Figure \ref{fig:confGraph} shows the confusion matrix in graphical form.  We can observe that while there are some pairs which are somewhat more commonly confused, overall the misclassifications are widely distributed. This means that while some authors may resemble each other, overall our low individual snippet accuracy is likely due to an overall lack of distinguishing information per snippet.

\subsection{Impact of Ensembling}

Simple averaging of probabilistic predictors has long been known to yield an
ensemble classifier that has significantly improved generalization performance
and robustness to error over any individual component
\cite{lincoln1990synergy}.  This improvement has also long been known to be
inversely related to the degree of correlation between the \emph{predictions} of
the classifiers on a single example \cite{tumer1996error}.

The standard approach for averaging considers an ensemble of learners
$h_1,\ldots h_T$, and takes the overall classification of a single sample $x$
to be:

\begin{equation*}
    H\left( x \right) = \frac{1}{T}\sum_{i=1}^{T}h_i(x)
\end{equation*}

We examine an interesting variation on this problem where, instead of
submitting a single test sample to a diverse set of classifiers, we submit a
diverse collection of test samples which are known to share the same unknown
label $x_1^{(i)}, x_2^{(i)}\ldots x_n^{(i)}$ to a single classifier, and
average their outputs to obtain a final classification:

\begin{equation*}
    H\left( i \right) = \frac{1}{n}\sum_{j=1}^{n}h\left( x_j^{(i)} \right)
\end{equation*}

The underlying intuition remains unchanged: if the erroneous components of any
given prediction are approximately uncorrelated between different code samples
from a single author (i.e., weight for incorrect predictions is approximately
uniformly distributed over those incorrect predictions, given a sufficient
number of samples), then in taking the average prediction across samples these
errors will cancel each other out, resulting in an improved prediction over any
of the individual samples.  

While full details are omitted due to space constraints, we evaluated this hypothesis in a 15-author data set by examining the coefficient of variation (the ratio of the standard deviation to the mean) across individual sample predictions across several test ensemble sizes (data not shown). Reliably, this variation is minimized for the correct prediction, indicating that from individual sample to individual sample within the ensemble, the predicted probability for the incorrect labels varies significantly relative to the mean, suggesting that they are in fact approximately uncorrelated. Significantly, when the coefficient of variation for the correct label is not the smallest, these samples tend to have the lowest ensemble confidence and also are most likely to be incorrectly predicted.  

\subsection{Ground Truth Considerations}
\label{ground-truth}

We acknowledge that ground truth in the GitHub environment is not perfect. First, we know that it is possible that code is copied from other sources, and therefore is not the original contribution of the credited author.  Furthermore, using git blame at the line level indicates the last author to touch the line, not the original author of all of the parts of the line.  We note that despite the inherent ``messiness'' of the data, it is important to evaluate on data collected from the wild which well reflects realistic application of our techniques.  Therefore, we devised an experiment to try to quantify the contribution of inaccurate ground truth.  

For this experiment, we perform cross validation on Google Code Jam data as is used by Caliskan-Islam et al.~\cite{caliskan2015anonymizing} with 250 authors and 9 files each.  We then set a parameter $m$ for number of ground truth corruptions to perform.  Each corruption is an author swap between two files.  We use Google Code Jam data because it gives us more reliable ground truth than GitHub.

\begin{figure}[tb]
\centering
\includegraphics[width=\columnwidth]{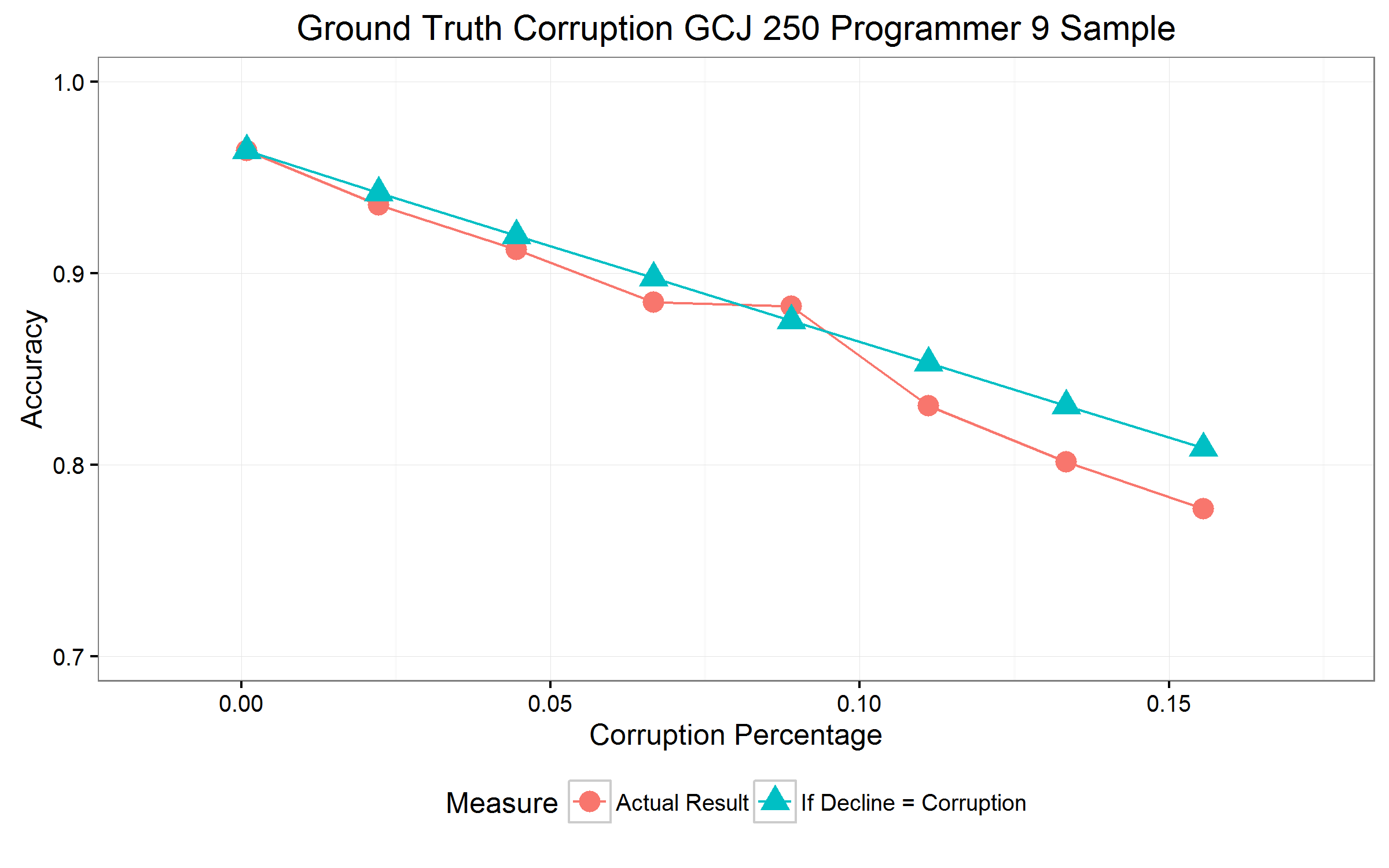}
\caption{These are the results for ground truth corruption in the Google Code Jam dataset.}
\label{fig:corruption}
\end{figure}

While random forests are known to be robust against mislabeled data due to using bootstrap sampling, which causes each sample to only affect some of the trees in the overall classifier, we nevertheless performed a brief empirical assessment of the potential impact of mislabeled training data. Figure \ref{fig:corruption} shows the accuracy for various levels of ground truth corruption for varying percentages of corrupted labels in the Google Code Jam dataset.  We observe that the magnitude of the decline in accuracy is close to the magnitude of the incorrect ground truth labels for relatively small amounts of corruption.  Therefore, we conclude that individual incorrect labels have only minimal effect on the overall quality of the classifier, and that it would take serious systemic ground truth problems to cause extreme classification problems.  

\section{Limitations}
\label{sec:limit}

While our study is the first observing the applicability of stylometric attribution to code fragments from collaborative code, we acknowledge that there are limitations to our study.  Although we scraped a large amount of GitHub data, we ended up with a relatively small dataset.  A large amount of this discard was due to concerns about having sufficient training data, which is a known problem with all stylometric analysis and a consequence of the laboratory conditions in which we collect training and evaluation data from the same source. A more thorough open world study which instead of discarding these samples treats them as part of the out of world data would be appropriate to fully quantify the effects of such loss, and a study into the actual availability of training data would be appropriate to clarify the overall applicability of source code stylometry.  We also note that while we excluded files which were entirely single authored, many files we used had a dominant author, allowing for some samples which, while not complete code files, were still very large. A study which examines attributability in different levels of collaboration would be useful.  Another limitation is that many authors only contributed to a single collaborative project. As a result, code samples are not from as different environments as we may expect in real application. A smaller scale study which enforces a project split for training and testing would help better understand the effects of this issue.

\section{Future Work}
\label{sec:future}

Our main results assume that we know that the correct programmer is one of our suspects and that we have a segmentation either in form of commits or git-blame. We have made strides in removing these assumptions, and would like to continue to do so.

In this work we present a way to remove the closed world assumption.  This assumption is common to stylometric work, but does not often match real world use cases.  Our technique allows easy elimination of most out of world samples at the cost of eliminating many correct, but difficult to trust, attributions in the case of single sample attribution, and some for account attribution.  While this technique provides a solid start towards addressing the open world problem, it would be preferable to find a technique which sacrifices fewer correctly attributed samples while improving our ability to trust the attributions of those difficult samples.

The primary assumption which remains can be considered the version control assumption.  While this can be a reasonable assumption, as most large collaborative projects are written under version control, we are not guaranteed access to the repository itself. 
In order to remove this assumption, we would need to perform source code segmentation which would split a large source code file into components by author.  Alternately, we could attempt to perform sliding window attribution on source code files. We are also interested in potentially performing segmentation by the AST instead of by lines, given that the most important features are AST-based. 

We further assume that commits and blames are similar types of data and that our results for blames would also be valid for commits.  However, it would be useful to confirm this, and if this hypothesis is false then we would like to find a technique allowing us to classify commits as well as blames.  Furthermore, we would like to use git-author as a best of both worlds scenario and to examine extensions of this work into multi-label classification for individual samples as well as whole files.

Because the accuracy for account attribution is superior to the accuracy for single sample attribution, we would like to find an unsupervised method to transform single sample attribution scenarios into scenarios closer to account attribution.

Additionally, the reliance of our classifier on highly specific features suggests that an ensemble classifier combining AST-based attribution for the general case and text attribution for highly specific samples may be worth considering, especially for real-world samples where comments may have important identifying information for some authors but not others. We would also be interested in investigating the performance of existing obfuscation tools such as Stunnix\cite{stunnix} and Tigress\cite{tigress} for anonymizing short source code samples.

As mentioned in our analysis, we also believe that we can create stronger predictors of attribution correctness than classifier confidence alone. One such indicator we suggest is to combine sample length with confidence, but even better predictors may be possible.

While we believe that there are important security applications for the ability to perform attribution at this level, we are also concerned about the serious privacy ramifications of this technology existing.  Therefore, we hope to develop techniques and tools to allow programmers to better anonymize themselves, and repository owners to facilitate anonymity for contributors.   We also hope to develop additional rules which indicate what makes a source code sample harder to attribute.

\section{Conclusion}
\label{sec:conclusion}

We show that it is possible to identify the author of even small, single code contributions, but easier to attribute collections of such contributions, such as those belonging to an account. We further show the joint use of confidence metrics and calibration curves can be used to perform attribution in the open world and in cases with lower overall accuracy. We observe through our analysis that it may be possible to use quantifiable higher level features of code - distinct from the low level machine learning features - to enhance the calibration curves to make even better distinctions between trustworthy and untrustworthy attributions. As a result, we suggest that analysts need to first perform a calibration step using labeled data in order to determine the best acceptance function for their purposes before attempting attribution of code fragments.

Our results show that attribution is easier if samples can be linked in advance, such as through a pseudonymous account on a version control system or other online platform. We have shown that having even two classification samples known to be by the same individual results in boosted accuracy, and adding more samples results in greater accuracy. We also observe that larger samples often, although not always, result in more trustworthy attributions. We have also observed that licensing and other pieces of code containing plain-text tend to be more easily identified, as is code containing highly personalized function declarations and macros - although these features could also be used to deceive classifiers. We have shown that some degree of anonymity for short code fragments may be possible, but not guaranteed, in the short term if precautions are taken, but our results also suggest that, as the state of the art in attribution advances and more features of the code are utilized, anonymity becomes harder to obtain.

\section{Acknowledgement}

We would like to thank DARPA Contract No. FA8750-17-C-0142
and the United States Army Research Laboratory Contract No. W911NF-15-2-0055 for funding this work.

\bibliographystyle{ACM-Reference-Format}
\bibliography{Main}

\begin{appendices}

\section{Sample Attributions}
\label{sec:example}

In this section, we show the entire attribution pipeline for selected example code snippets. For these examples, we will examine a single train-test split in the closed world. Each of the code samples in this section belong to author 74 from our dataset, and for simplicity will be named sample A, B, C, D, E, and F respectively. Samples A and B were taken from one file and samples C through F were taken from another. The samples are shown in \Cref{fig:sampA,fig:sampB,fig:sampC,fig:sampD,fig:sampE,fig:sampF}, with line breaks added and removed for formatting reasons. As stated previously, all samples are artificially enclosed in a dummy main function in order to enable parsing with joern \cite{YamGolArpRie14}.

\begin{figure}
\begin{verbatim}
int main(){
  exp2_ = n;
  num_ = 1 << n;
}
\end{verbatim}
\caption{Code Sample A}
\label{fig:sampA}
\end{figure}

\begin{figure}
\begin{verbatim}
int main(){
    buf_ = new float[num_];
    tab_ = new float[num_ * 2];
}
\end{verbatim}
\caption{Code Sample B}
\label{fig:sampB}
\end{figure}

\begin{figure}
\begin{verbatim}
int main(){
  const double F =
      static_cast<double>(height() - 2) /
      (log10(255) * MAX_AMPLITUDE);
}
\end{verbatim}
\caption{Code Sample C}
\label{fig:sampC}
\end{figure}

\begin{figure}
\begin{verbatim}
int main(){
  updateBandSize(band_count_);
  colorChanged();
  setMinimumSize(QSize(band_count_ *
  kColumnWidth, 10));
}
}
\end{verbatim}
\caption{Code Sample D}
\label{fig:sampD}
\end{figure}

\begin{figure}
\begin{verbatim}
int main(){
void BarAnalyzer::colorChanged() {
  if (pixBarGradient_.isNull()) {
    return;
  }
  QPainter p(&pixBarGradient_);
  QColor rgb;
  if (psychedelic_enabled_) {
    rgb = getPsychedelicColor(scope_, 50, 100);
  } else {
    rgb = palette().color(QPalette::Highlight);
  }
  for (int x = 0, r = rgb.red(), 
  		g = rgb.green(),
        b = rgb.blue(), r2 = 255 - r; 
        x < height(); ++x) {
}
\end{verbatim}
\caption{Code Sample E}
\label{fig:sampE}
\end{figure}

\begin{figure}
\begin{verbatim}
int main(){
    if (roofVelocityVector_[i] != 0) {
      if (roofVelocityVector_[i] > 32)  
      // no reason to do == 32
        roofVector_[i] -=
            (roofVelocityVector_[i] - 32) / 20;  
            // trivial calculation
}
\end{verbatim}
\caption{Code Sample F}
\label{fig:sampF}
\end{figure}

\begin{figure}[tb]
\centering
\includegraphics[width=\columnwidth]{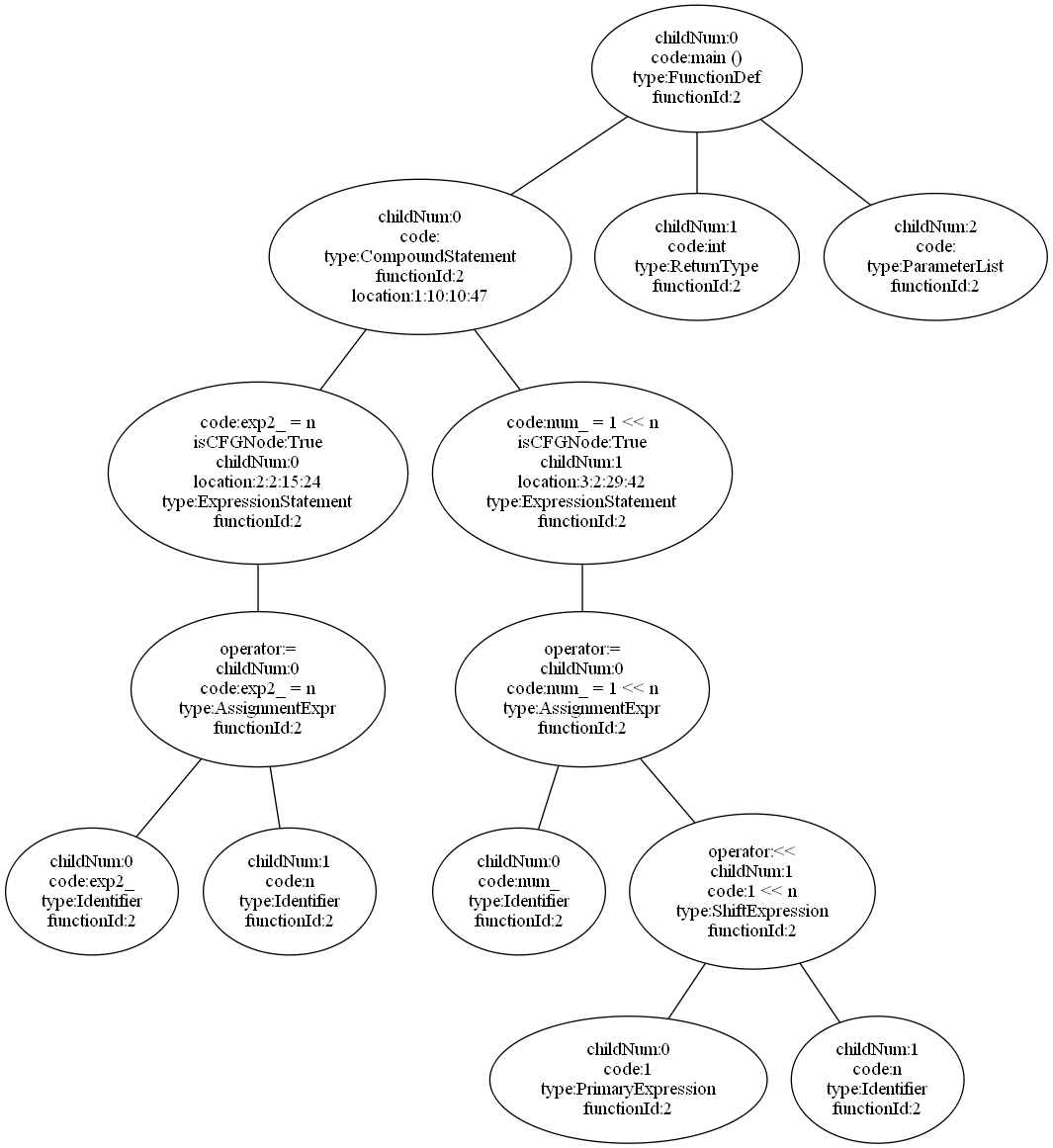}
\caption{This is the AST for sample A produced by joern.}
\label{fig:samAast}
\end{figure}

After identifying our samples, the next step is to extract the abstract syntax trees. Figure \ref{fig:samAast} shows the extracted abstract syntax tree for sample A.  Once we extract ASTs we need to transform those into features, along with text from the code. For example, sample F would have a word unigram feature representing the word ``reason'' while sample A would have a feature corresponding to the shift expression node, as well as a feature representing the depth of the node in the tree.

Table \ref{tab:samRes} shows the classification results for each sample, indicating the predicted class, the confidence for the predicted class, and the confidence for the true class. If we refer back to our calibration curve, we'd say that sample A has about 30\% chance of being by author 42, sample B has about 30\% chance of being by author 44, sample F has about 30\% chance of being by author 51, samples D and E have about 30\% chance of being by author 74, and sample C has about 55\% chance of being by author 74.

We can then take varying combinations of these samples for account attribution. \Cref{tab:pairRes,tab:trioRes,tab:quadRes,tab:quinRes} show the results for the various combinations of samples in accounts. If we take all of the samples together, we receive a prediction of author 74 with confidence 6.5\%. Refering back to the calibration curve, most of these predictions have about a 55\% chance of being accurate, with those with over 10\% confidence having between a 55\% and 80\% chance of being accurate.  However, if we consider the predictions of all of the possible combinations we can be slightly more optimistic in our prediction accuracy.  Further, it is reasonable to assume that if we had a more granular calibration curve we would also evaluate our chance of success more highly.

\begin{table}[]
\centering
\caption{Example classification confidence results}
\label{tab:samRes}
\begin{tabular}{|l|l|l|l|}
\hline 
Sample & Prediction & P Confidence & T Confidence\tabularnewline
\hline 
\hline 
A & 42 & 9.8\% & 0.4\%\tabularnewline
\hline
B & 44 & 8.2\% & 0.6\%\tabularnewline
\hline
C & 74 & 19.6\% & 19.6\%\tabularnewline
\hline
D & 74 & 6.4\% & 6.4\%\tabularnewline
\hline
E & 74 & 7.6\% & 7.6\%\tabularnewline
\hline
F & 51 & 5.8\% & 4.4\%\tabularnewline
\hline
\end{tabular}
\end{table}

\begin{table}[]
\centering
\caption{Example pairs classification confidence}
\label{tab:pairRes}
\begin{tabular}{|l|l|l|l|}
\hline 
Samples & Prediction & P Confidence & T Confidence\tabularnewline
\hline 
\hline 
AB & 44 & 5.7\% & 0.5\%\tabularnewline
\hline
AC & 74 & 10.0\% & 10.0\%\tabularnewline
\hline
AD & 42 & 5.7\% & 3.4\%\tabularnewline
\hline
AE & 42 & 4.9\% & 4.0\%\tabularnewline
\hline
AF & 46 & 5.0\% & 2.4\%\tabularnewline
\hline
BC & 74 & 10.1\% & 10.1\%\tabularnewline
\hline
BD & 44 & 4.2\% & 3.5\%\tabularnewline
\hline
BE & 44 & 4.5\% & 4.1\%\tabularnewline
\hline
BF & 44 & 4.4\% & 2.5\%\tabularnewline
\hline
CD & 74 & 13.0\% & 13.0\%\tabularnewline
\hline
CE & 74 & 13.6\% & 13.6\%\tabularnewline
\hline
CF & 74 & 12.0\% & 12.0\%\tabularnewline
\hline
DE & 74 & 7.0\% & 7.0\%\tabularnewline
\hline
DF & 74 & 5.4\% & 5.4\%\tabularnewline
\hline
EF & 74 & 6.0\% & 6.0\%\tabularnewline
\hline
\end{tabular}
\end{table}

\begin{table}[]
\centering
\caption{Example triples classification confidence}
\label{tab:trioRes}
\begin{tabular}{|l|l|l|l|}
\hline 
Samples & Prediction & P Confidence & T Confidence\tabularnewline
\hline 
\hline 
ABC & 74 & 6.9\% & 6.9\%\tabularnewline
\hline
ABD & 69 & 4.1\% & 2.5\%\tabularnewline
\hline
ABE & 44 & 4.1\% & 2.9\%\tabularnewline
\hline
ABF & 46 & 4.3\% & 1.8\%\tabularnewline
\hline
ACD & 74 & 8.8\% & 8.8\%\tabularnewline
\hline
ACE & 74 & 9.2\% & 9.2\%\tabularnewline
\hline
ACF & 74 & 8.1\% & 8.1\%\tabularnewline
\hline
ADE & 74 & 4.8\% & 4.8\%\tabularnewline
\hline
ADF & 46 & 4.0\% & 3.7\%\tabularnewline
\hline
AEF & 74 & 4.1\% & 4.1\%\tabularnewline
\hline
BCD & 74 & 8.9\% & 8.9\%\tabularnewline
\hline
BCE & 74 & 9.3\% & 9.3\%\tabularnewline
\hline
BDE & 74 & 8.2\% & 8.2\%\tabularnewline
\hline
BDF & 74 & 4.9\% & 4.9\%\tabularnewline
\hline
BEF & 74 & 3.8\% & 3.8\%\tabularnewline
\hline
CDE & 74 & 4.2\% & 4.2\%\tabularnewline
\hline
CDF & 74 & 11.2\% & 11.2\%\tabularnewline
\hline
CEF & 74 & 10.1\% & 10.1\%\tabularnewline
\hline
DEF & 74 & 6.1\% & 6.1\%\tabularnewline
\hline
\end{tabular}
\end{table}

\begin{table}[]
\centering
\caption{Example quadruples classification confidence}
\label{tab:quadRes}
\begin{tabular}{|l|l|l|l|}
\hline 
Samples & Prediction & P Confidence & T Confidence\tabularnewline
\hline 
\hline 
ABCD & 74 & 6.8\% & 6.8\%\tabularnewline
\hline
ABCE & 74 & 7.1\% & 7.1\%\tabularnewline
\hline
ABCF & 74 & 6.1\% & 6.2\%\tabularnewline
\hline
ABDE & 74 & 3.8\% & 3.8\%\tabularnewline
\hline
ABDF & 46 & 3.8\% & 3.0\%\tabularnewline
\hline
ABEF & 46 & 3.6\% & 3.3\%\tabularnewline
\hline
ACDE & 74 & 8.5\% & 8.5\%\tabularnewline
\hline
ACDF & 74 & 7.7\% & 7.7\%\tabularnewline
\hline
ACEF & 74 & 8.0\% & 8.0\%\tabularnewline
\hline
ADEF & 74 & 4.7\% & 4.7\%\tabularnewline
\hline
BCDE & 74 & 8.6\% & 8.6\%\tabularnewline
\hline
BCDF & 74 & 7.8\% & 7.8\%\tabularnewline
\hline
BCEF & 74 & 8.1\% & 8.1\%\tabularnewline
\hline
BDEF & 74 & 4.8\% & 4.8\%\tabularnewline
\hline
CDEF & 74 & 9.5\% & 9.5\%\tabularnewline
\hline
\end{tabular}
\end{table}

\begin{table}[]
\centering
\caption{Example quintuples classification confidence}
\label{tab:quinRes}
\begin{tabular}{|l|l|l|l|}
\hline 
Samples & Prediction & P Confidence & T Confidence\tabularnewline
\hline 
\hline 
ABCDE & 74 & 6.9\% & 6.9\%\tabularnewline
\hline
ABCDF & 74 & 6.3\% & 6.3\%\tabularnewline
\hline
ABCEF & 74 & 6.5\% & 6.5\%\tabularnewline
\hline
ABDEF & 74 & 3.9\% & 3.9\%\tabularnewline
\hline
ACDEF & 74 & 7.7\% & 7.7\%\tabularnewline
\hline
BCDEF & 74 & 7.7\% & 7.7\%\tabularnewline
\hline
\end{tabular}
\end{table}

\section{Relaxed Attribution}
\label{sec:relax}

\begin{figure}[htb]
\centering
\includegraphics[width=\columnwidth]{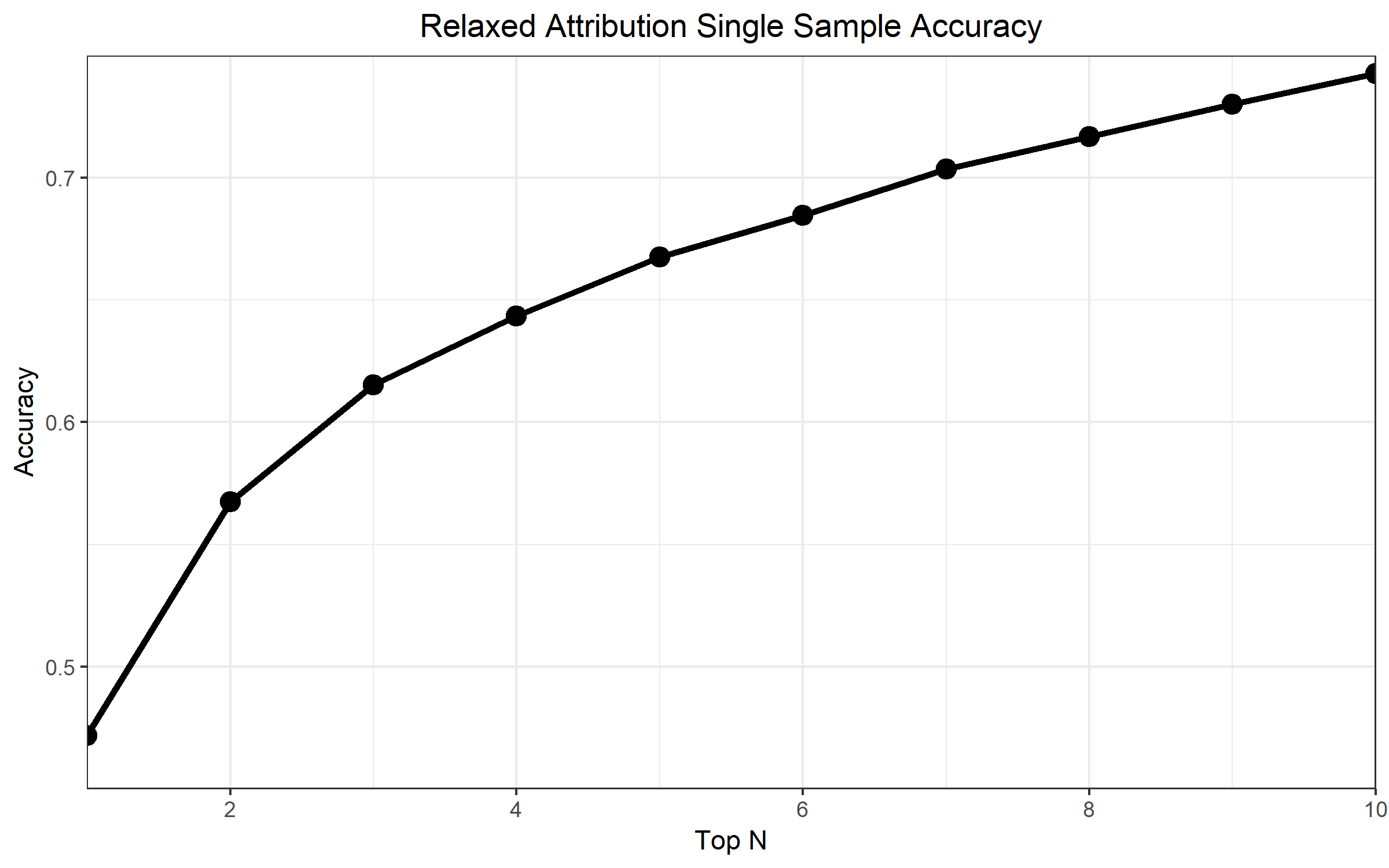}
\caption{Accuracy increases as the number $N$ of top authors to accept increases.}
\label{fig:relax}
\end{figure}

In addition to other analyses, we consider relaxed attribution to see to what extent we can reduce the suspect set for more difficult samples. Figure \ref{fig:relax} shows the results for relaxed attribution for single sample attribution only.  In the event of a tie, we consider all tied elements on the $N$th place as equal. We show that it is possible to reduce the suspect set to approximately one tenth with nearly 75\% accuracy.

\end{appendices}

\end{document}